%% file: root.tex
\documentclass[10pt,twocolumn,letterpaper]{article}

\newif\ifarxiv
\newif\ifcvpr
\newif\ifcvprfinal

\cvprfalse \cvprfinalfalse \arxivtrue %

\ifcvprfinal
\usepackage{cvpr} %
\fi
\ifcvpr
\usepackage[review]{cvpr} %
\fi
\ifarxiv
\usepackage[pagenumbers]{cvpr} %
\fi

\input{includes/packages}

\input{includes/colors_and_icons}

\input{includes/macros}

\newcommand{\coolname}{Map\-Anything\xspace}

\newcommand{\webpagetext}{map-anything.github.io}
\newcommand{\webpage}{https://map-anything.github.io/}

\def\paperID{235} %
\def\confName{3DV}
\def\confYear{2026}

\ifcvprfinal
\title{\coolname: Universal Feed-Forward Metric 3D Reconstruction \\[1.5pt]
\Large{\href{\webpage}{\color{darkorange}{\webpagetext}}}
\vspace{-1.2em}
}
\fi
\ifcvpr
\title{\coolname: Universal Feed-Forward Metric 3D Reconstruction}
\fi
\ifarxiv
\title{\coolname: Universal Feed-Forward Metric 3D Reconstruction \\[1.6pt]
\Large{\href{\webpage}{\color{darkorange}{\webpagetext}}}
\vspace{-1em}
}
\fi

\newcommand{\supptitle}[1]{
\coolname: Universal Feed-Forward Metric 3D Reconstruction
\vspace{-0.4em}
}

\definecolor{citegray}{gray}{0.32}
\newcommand{\authorhref}[3][citegray]{\href{#2}{\color{#1}{#3}}}
\def\authorspace{\hspace{.1in}}

\author{%
\authorhref{https://nik-v9.github.io/}{\textbf{Nikhil Keetha}}\hspace{.02in}$^{1, 2}$
\authorspace
\authorhref{https://sirwyver.github.io/}{\textbf{Norman Müller}}\hspace{.02in}$^{1}$
\authorspace
\authorhref{https://demuc.de/}{\textbf{Johannes Schönberger}}\hspace{.02in}$^{1}$
\authorspace
\authorhref{https://www.linkedin.com/in/lorenzoporzi}{\textbf{Lorenzo Porzi}}\hspace{.02in}$^{1}$
\authorspace
\authorhref{https://infinity1096.github.io/}{\textbf{Yuchen Zhang}}\hspace{.02in}$^{2}$
\\[0.25em]
\authorhref{https://tobiasfshr.github.io/}{\textbf{Tobias Fischer}}\hspace{.02in}$^{1}$
\authorspace
\authorhref{https://www.linkedin.com/in/arno-knapitsch}{\textbf{Arno Knapitsch}}\hspace{.02in}$^{1}$
\authorspace
\authorhref{https://www.linkedin.com/in/duncan-zauss}{\textbf{Duncan Zauss}}\hspace{.02in}$^{1}$
\authorspace
\authorhref{https://ethanweber.me/}{\textbf{Ethan Weber}}\hspace{.02in}$^{1}$
\authorspace
\authorhref{https://www.linkedin.com/in/nelsonantunes7}{\textbf{Nelson Antunes}}\hspace{.02in}$^{1}$
\\[0.25em]
\authorhref{https://x.com/jonathonluiten?lang=en}{\textbf{Jonathon Luiten}}\hspace{.02in}$^{1}$
\authorspace
\authorhref{https://m.lopezantequera.com/}{\textbf{Manuel Lopez-Antequera}}\hspace{.02in}$^{1}$
\authorspace
\authorhref{https://scholar.google.com/citations?user=484sccEAAAAJ}{\textbf{Samuel Rota Bul\`{o}}}\hspace{.02in}$^{1}$
\authorspace
\authorhref{https://richardt.name/}{\textbf{Christian Richardt}}\hspace{.02in}$^{1}$
\\[0.25em]
\authorhref{https://www.cs.cmu.edu/~deva/}{\textbf{Deva Ramanan}}\hspace{.02in}$^{2}$
\authorspace
\authorhref{https://theairlab.org/team/sebastian/}{\textbf{Sebastian Scherer}}\hspace{.02in}$^{2}$
\authorspace
\authorhref{https://www.linkedin.com/in/peter-kontschieder-2a6410134}{\textbf{Peter Kontschieder}}\hspace{.02in}$^{1}$
\\[0.65em]
$^{1}$\hspace{.02in}\href{https://www.meta.com/}{\textcolor{citegray}{\textbf{Meta Reality Labs}}}
\hspace{1em}
$^{2}$\hspace{.02in}\href{https://www.ri.cmu.edu/}{\textcolor{citegray}{\textbf{Carnegie Mellon University}}}%
}

\begin{document}

\twocolumn[{
\maketitle
    \centering
    \vspace{-0.25in}
    \includegraphics[trim={0cm 6cm 0cm 0cm},clip,width=\textwidth]{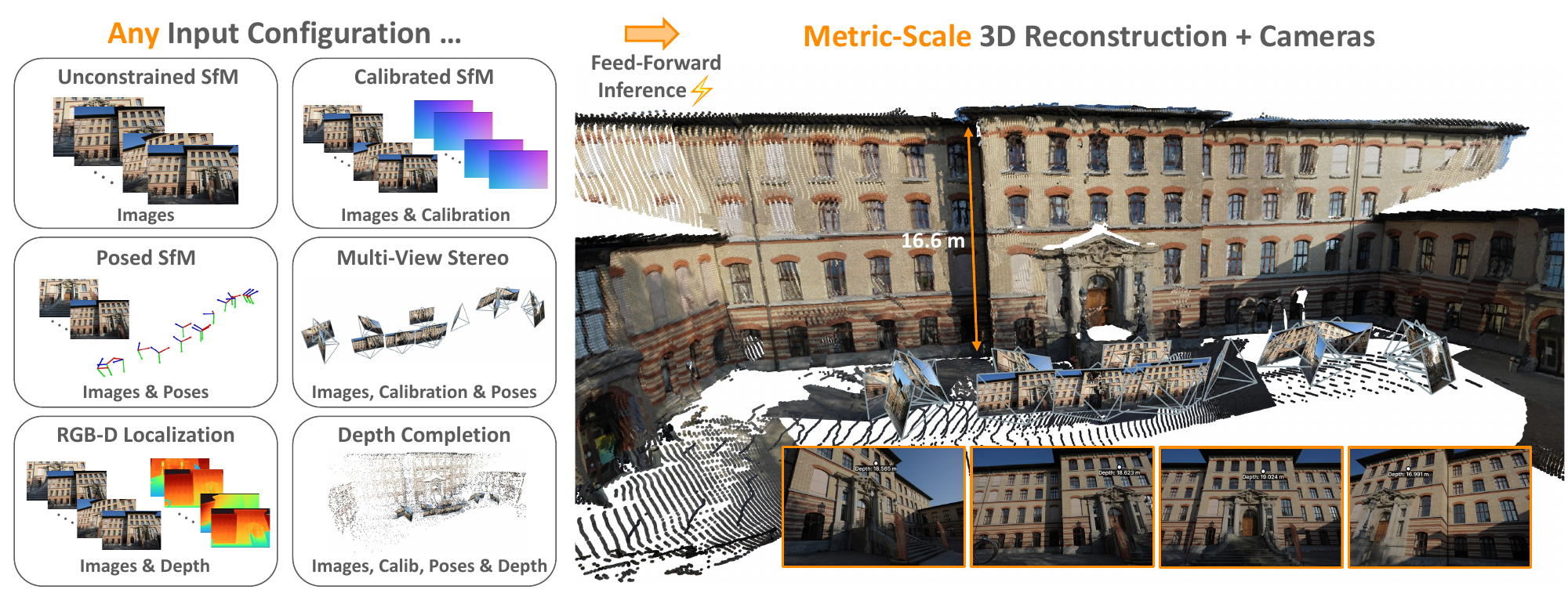}
    \captionsetup{hypcap=false}
    \captionof{figure}{%
    \textbf{\coolname is a flexible, unified feed-forward 3D reconstruction model} that predicts metric 3D reconstructions with camera information from a set of $N$ input images with optional camera poses, intrinsics, or depth maps.
    \coolname supports over 12 different 3D reconstruction tasks, including camera localization, structure-from-motion (SfM), multi-view stereo, and metric depth completion, outperforming or matching the quality of specialist methods. 
    }
    \label{fig:splash}
    \vspace{1em}
}] 

\input{text/00_abstract}
\input{text/01_intro}

\input{text/02_related}
\input{text/03_method}

\input{text/04_benchmarking}

\input{text/05_conclusion}

\ifarxiv
\input{supplementary}
\fi

{
    \small
    \bibliographystyle{ieeenat_fullname}
    \bibliography{root, MapAnything-CR}
}

\ifcvpr
\input{supplementary}
\fi
\ifcvprfinal
\input{supplementary}
\fi

\end{document}

%% file: includes/colors_and_icons.tex
\definecolor{Gray}{gray}{0.50}
\newcolumntype{g}{>{\columncolor{Gray}}c}
\definecolor{ffe1da}{RGB}{255,225,218}
\definecolor{F7E0D5}{RGB}{247,224,213}
\definecolor{darkF7E0D5}{RGB}{209,154,128}
\colorlet{Light}{White!0!F7E0D5}
\definecolor{darkorange}{rgb}{1.0, 0.54, 0}

\colorlet{tabfirst}{Green!25}
\definecolor{tabthird}{rgb}{1, 0.85, 0.7}
\definecolor{tabsecond}{rgb}{1, 0.96, 0.7}

%% file: includes/macros.tex
\definecolor{darkorange}{rgb}{1.0, 0.54, 0}
\definecolor{darkpurple}{rgb}{0.288,0.1196,0.7}

\definecolor{amber}{rgb}{1.0, 0.75, 0.0}

\makeatletter
\DeclareRobustCommand\onedot{\futurelet\@let@token\@onedot}
\def\@onedot{\ifx\@let@token.\else.\null\fi\xspace}

\def\ie{i.e\onedot}

\renewcommand\paragraph{\@startsection{paragraph}{4}{\z@}%
  {4pt}%
  {-12pt}%
  {\normalfont\normalsize\bfseries}}
\makeatother

\expandafter\def\expandafter\normalsize\expandafter{%
    \normalsize%
    \setlength\abovedisplayskip{0pt plus 2pt}%
    \setlength\belowdisplayskip{0pt plus 2pt}%
    \setlength\abovedisplayshortskip{0pt plus 2pt}%
    \setlength\belowdisplayshortskip{0pt plus 2pt}%
}

\newcommand{\Rom}[1]{\uppercase\expandafter{\romannumeral #1\relax}}
\newcommand{\R}{\mathbb{R}}
\newcommand{\abs}[1]{\left| #1 \right|}

\newcommand{\metric}[1]{{#1}^\text{metric}}
\newcommand{\uts}[1]{\tilde{#1}}  %

\definecolor{darkgray}{rgb}{0.2, 0.2, 0.2}

\usepackage[capitalize,nameinlink,noabbrev]{cleveref}
\crefformat{footnote}{#2\footnotemark[#1]#3}



\newcommand*{\subfigref}[2][]{%
  Fig. \hyperref[{fig:#2}]{%
    \ref*{fig:#2}%
    \ifx\\#1\\%
    \else
      \,#1%
    \fi
  }%
}

\makeatletter

\newcommand*{\@rowstyle}{}

\newcommand*{\rowstyle}[1]{%
  \gdef\@rowstyle{#1}%
  \@rowstyle\ignorespaces%
}

\newcolumntype{=}{%
  >{\gdef\@rowstyle{}}%
}

\newcolumntype{+}{%
  >{\@rowstyle}%
}

\makeatother

\newcommand{\absrel}{\textrm{rel}}
\newcommand{\threshI}{\tau}

\newcommand{\bestresult}[1]{\textbf{#1}}

\definecolor{darkgreen}{RGB}{0, 100, 0}
\newcommand{\greencheck}{{\color{darkgreen}\checkmark}}
\newcommand{\redx}{{\color{red}\ding{55}}}

\newcommand{\tickmark}{\ding{51}}
\newcommand{\xmark}{\ding{55}}

%% file: text/00_abstract.tex
\begin{abstract}
We introduce MapAnything, a unified transformer-based feed-forward model that ingests one or more images along with optional geometric inputs such as camera intrinsics, poses, depth, or partial reconstructions, and directly regresses the metric 3D scene geometry and cameras.
MapAnything leverages a factored representation of multi-view scene geometry, \ie, a collection of depth maps, local raymaps, camera poses, and a metric scale factor that effectively upgrades local reconstructions into a globally consistent metric frame.
Standardizing the supervision and training across diverse datasets, along with flexible input augmentation, enables MapAnything to address a broad range of 3D vision tasks in a single feed-forward pass, including uncalibrated structure-from-motion, calibrated multi-view stereo, monocular depth estimation, camera localization, depth completion, and more.
We provide extensive experimental analyses and model ablations demonstrating that MapAnything outperforms or matches specialist feed-forward models while offering more efficient joint training behavior, thus paving the way toward a universal 3D reconstruction backbone.

\end{abstract}

%% file: text/01_intro.tex
\vspace{-1em}
\section{Introduction}
\label{sec:intro}

The problem of image-based 3D reconstruction has traditionally been solved using
structure-from-motion (SfM) \cite{SchoenF2016, PanBPS2024},
photometric stereo \cite{Woodh1980},
shape-from-shading \cite{horn1989obtaining},
and so on.
To make the problem tractable, classic approaches
decompose it into distinct tasks, such as
feature detection \cite{Lowe2004} and matching \cite{SarliDMR2020},
two-view pose estimation \cite{Niste2004},
camera calibration \cite{VeichSLP2024} and resectioning \cite{SattlLK2011},
rotation \cite{HartlTDL2013} and translation averaging \cite{PanBPS2024},
bundle adjustment (BA) \cite{TriggMHF2000}, multi-view stereo (MVS) \cite{SchoenZFP2016},
and/or monocular surface estimation \cite{HoiemEH2005a}.
Recent work has demonstrated tremendous potential in solving these problems in a unified way using feed-forward architectures \cite{WangLCCR2024, LeroyCR2024, ZhangLKYRT2024, WangTBXLSWXZ2024, IzquiSFGTCABW2025, DongWLCFKY2025}.

While prior feed-forward work has approached the different tasks separately or by not leveraging all the available input modalities, we present a unified end-to-end model for diverse 3D reconstruction tasks.
Our method \coolname can be used to solve the most general uncalibrated SfM problem as well as various combinations of sub-problems, such as calibrated SfM or multi-view stereo, monocular depth estimation, camera localization, and metric depth completion.
To enable the training of such a unified model, we:
(1) introduce a flexible input scheme that supports various geometric modalities when available,
(2) propose a suitable output space that supports all of these diverse tasks, and
(3) discuss flexible dataset aggregation and standardization.

\coolname's key insight to address these challenges is the use of a \emph{factored} representation of multi-view scene geometry.
Instead of directly representing the scene as a collection of pointmaps, we represent the scene as a collection of depth maps, local raymaps, camera poses, and a metric scale factor that upgrade local reconstructions into a globally consistent metric frame.
We use such a factored representation to represent both the outputs and (optional) inputs for \coolname, allowing it to take advantage of auxiliary geometric inputs when available.
For example, robotic applications \cite{HuXJFPKKXZFZOKASJBWSWKXB2023, KeethKJYSRL2024, ho2024map, AlamaBHKQWHKS2025} may have knowledge of camera intrinsics (rays) and/or extrinsics (poses).
Finally, a significant benefit of our factored representation is that it allows \coolname to be effectively trained from diverse datasets with partial annotations, for example, datasets that may be annotated with only non-metric ``up-to-scale'' geometry.
In summary, we make the following main contributions:
\begin{enumerate}
    \item \textbf{Unified Feed-Forward Model}
    for multi-view metric 3D reconstruction that supports more than 12 different problem configurations.
    The end-to-end transformer is trained more efficiently than a naive set of bespoke models and leverages not only image inputs, but also optional geometric information such as camera intrinsics, extrinsics, depth, and/or metric scale factor, when available.

    \vspace{0.6em}

    \item \textbf{Factored Scene Representation}
    that flexibly enables decoupled inputs and effective prediction of metric 3D reconstructions.
    Our model computes multi-view pixel-wise scene geometry and cameras directly, without redundancies or costly post-processing. 

    \vspace{0.6em}

    \item \textbf{State-of-the-Art Performance}
    compared to other feed-forward models, matching or surpassing expert models that are tailored for specific, isolated tasks.

    \vspace{0.6em}
    
    \item \textbf{Open Source Release}
    of (a) code for data processing, inference, benchmarking, training \& ablations, and (b) a pre-trained \coolname model under the permissive Apache 2.0 license, thereby providing an extensible \& modular framework plus model to facilitate future research on building 3D/4D foundation models.

\end{enumerate}

%% file: text/02_related.tex
\section{Related Work}
\label{sec:related}

\paragraph{Towards Universal 3D Reconstruction.}

In contrast to the traditional approach of designing specialized methods for distinct reconstruction tasks, recent efforts have shown great promise in solving them jointly with a single feed-forward architecture.
Early works like DeMoN \cite{UmmenZUMIDB2017}, DeepTAM \cite{ZhouUB2020} or DeepV2D \cite{TeedD2020} explored this direction with CNNs but did not match the performance of classical expert models.
Enabled by advances in deep learning, recent methods like PF-LRM \cite{WangTBXLSWXZ2024}, RayDiffusion \cite{ZhangLKYRT2024}, DUSt3R \cite{WangLCCR2024}, VGGSfM \cite{WangKRN2024}, and VGGT \cite{WangCKVRN2025} scale up transformers on large amounts of data.
Despite this breakthrough, these methods are still limited to
a subset of 3D reconstruction tasks with fixed inputs and output modalities, %
a small or fixed number of views, or they only work well in relatively constrained, typically object-centric, scenarios.
With \coolname, we overcome these limitations by designing a geometrically grounded and flexible architecture that supports heterogeneous input and output modalities for any number of input views.

\begin{figure*}[!t]
\centering
\includegraphics[width=0.9\textwidth]{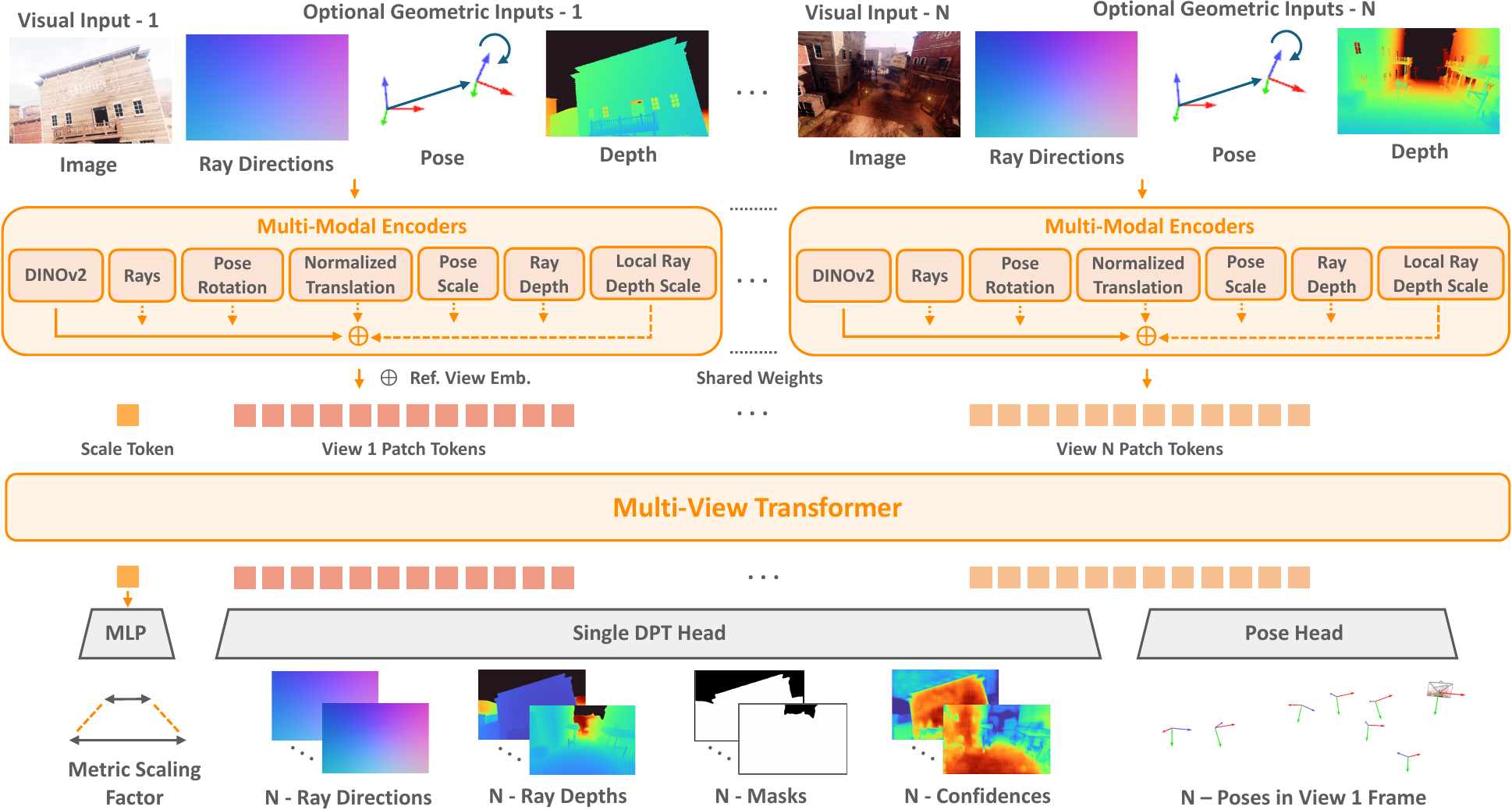}
\caption{\label{fig:method}%
\textbf{Overview of the \coolname Architecture.}
Given $N$ visual and optional geometric inputs, the model first encodes the images and the factored representation of the geometric inputs into a common latent space where the patch features (for images, rays \& depth) and broadcasted global features (for translation, rotation, pose scale across all pose inputs \& depth scale local to each frame) are summed together.
Then, a fixed reference view embedding is added to the first view's features and a single learnable scale token is appended to the set of $N$ view patch tokens.
These tokens are then input into an alternating-attention transformer.
We use a single DPT to decode the $N$ view patch tokens into $N$ dense outputs local to all the views.
A single average pooling-based pose head also uses the $N$ view patch tokens to predict $N$ poses in the frame of view 1.
Lastly, while these predictions exist in an up-to-scale space, the model passes the scale token through an MLP to predict the metric scaling factor, which when coupled with the other predictions, provides the dense metric 3D reconstruction.
}
\end{figure*}

\paragraph{Multi-View Feed-forward Reconstruction.}

DUSt3R and its metric follow-up MASt3R \cite{LeroyCR2024} predict a coupled scene representation (\ie, cameras, poses, and geometry are parameterized by a pointmap and need to be recovered post hoc) and require expensive post-processing \& symmetric inference to perform multi-view unconstrained SfM.
Follow-up work \cite{DuistZWLCR2025, MuraiDD2025, ElfleZAL2025, PatakSSP2025} integrates MASt3R outputs into classical SfM and SLAM pipelines in a more principled manner.
Recent works like Spann3R \cite{WangA2025}, CUT3R \cite{WangZHEK2025}, and MUSt3R \cite{CabonSACCRL2025} remove the need for classical optimization and enable multi-view reconstruction via latent state memory in transformers.
However, these works do not yet match the performance of traditional optimization applied to predicted two-view outputs from MASt3R \cite{DuistZWLCR2025,MuraiDD2025}.

Recently, MV-DUSt3R+ \cite{TangFWXRSY2025} and VGGT \cite{WangCKVRN2025} demonstrate multi-view inference by extending the DUSt3R architecture for multi-view reconstruction.
Likewise, Reloc3r \cite{DongWLCFKY2025} focuses on camera re-localization and directly predicts multi-view camera poses.
MV-DUSt3R+ achieves this by parallelizing the cross-attention transformer to support different reference views, leading to a significant increase in computation, while VGGT employs an alternating attention transformer to predict multi-view pointmaps, depth, pose, and features for tracking.
FASt3R \cite{YangSLHTCCMF2025} uses positional encoding for long-sequence inference in LLMs for global attention trained on a few views to work on a larger number of views.
More recently, $\pi^3$~\cite{WangZZCZLCPSH2025} fine-tunes VGGT to remove the use of the first input frame as reference coordinate.

In both MV-DUSt3R+ and FASt3R, the prediction is a coupled scene representation and cannot handle heterogeneous inputs.
As shown in FASt3R, for the multi-view setup, the dense geometry prediction capabilities of the model are impacted by the pose estimation across non-visible views (see their Table 5 \& Section 5.1).
To alleviate this issue, FASt3R predicts redundant pointmaps across all views with a dedicated DPT head for global and local pointmap prediction.
Likewise, VGGT also predicts multiple redundant quantities through two separate branches,
one for pointmaps and one for cameras and depth.
While concurrent work, $\pi^3$~\cite{WangZZCZLCPSH2025}, fine-tunes VGGT to remove this redundancy by predicting up-to-scale decoupled local pointmaps and global pose, we find this design choice to be sub-optimal (see \cref{tab:ablation_rep}).
In contrast, \coolname directly predicts a completely factored representation, i.e., local ray directions, depth along the ray, global camera pose for all views, and a single metric scaling factor for the scene.
In this formulation, the task of predicting ray directions (akin to camera calibration) and depth-along-ray estimation are per-view and thus can be predicted from a single dense prediction head.

While prior work has paved the way for unconstrained multi-view inference and large-scale training, they are all limited to only image inputs and modeling a simple pinhole camera. %
In contrast, \coolname supports various 3D reconstruction and calibration tasks from multiple views with heterogeneous inputs and a flexible camera model.

\paragraph{Geometry as Inputs or Conditioning.}

While not explicitly used for feed-forward 3D reconstruction like in \coolname,
quantities such as ray directions, origins, and depth maps have been explored as conditioning inputs for tasks like novel-view synthesis
\cite{JinJTZBZLSX2025, LuZFNTQCYL2025, WeberMKAZKR2026, ZhouGVVYBTRJ2025}, 
diffusion-based image generation
\cite{MouWXWZQSQ2024, ZhangRA2023}, 
dynamic video depth estimation \cite{LuHLDLCDYWL2025}, or
3D object shape completion \cite{DuistOWBRI2025}.
Taskonomy \cite{ZamirSSGMS2018} explored
the benefits of multi-task learning for improved vision task performance.
Later works like MultiMAE \cite{BachmMAZ2022} build on these insights and devise auto-encoders to support flexible combination of heterogeneous inputs; however, this is not suitable for solving 3D reconstruction tasks.
Pow3R \cite{JangWLAR2025} was the first work to incorporate known priors as inputs to feed-forward 3D reconstruction.
In contrast to us, Pow3R only supports two pinhole camera images with a single focal length and centered principal point.
Furthermore, Pow3R builds on top of DUSt3R and cannot condition on metric scale information.
In contrast, \coolname supports any number of input views and has a flexible input parameterization that supports metric scale and any camera with a central projection model.

%% file: text/03_method.tex
\section{\coolname{}}
\label{sec:approach}

\label{subsec:approach_overview}

\coolname is an end-to-end model that takes as input $N$ RGB images $\hat{\mathcal{I}} \!=\! (\hat{I}_i)_{i=1}^N$ and optional geometric inputs corresponding to all or a subset of the input views:
\begin{enumerate}[label=(\textbf{\alph*})]
\item %
generic central camera calibrations \cite{GrossN2001,VasilGAPBSG2020,ZhangLKYRT2024} as ray directions $\hat{\mathcal{R}} \!=\! (\hat{R}_i)_{i \in S_\text{r}}$,

\item %
poses in the frame of the first view $\hat{I}_1$ as quaternions $\hat{\mathcal{Q}} \!=\! (\hat{Q}_i)_{i \in S_\text{q}}$ and translations $\hat{\mathcal{T}} \!=\! (\hat{T}_i)_{i \in S_\text{t}}$, and

\item %
ray depth for each pixel $\hat{\mathcal{D}} \!=\! (\hat{D}_i)_{i \in S_\text{d}}$,
\end{enumerate}
where $S_\text{r}, S_\text{q}, S_\text{t}, S_\text{d}$ are subsets of frame indices $[1, N]$.

\coolname maps these inputs to an $N$-view factored metric 3D output (as shown in \cref{fig:method}):
\begin{equation}
    \label{eq:MapAnything}
    \!f_\text{MapAnything}\bigl(\hat{\mathcal{I}}, [\hat{\mathcal{R}}, \hat{\mathcal{Q}}, \hat{\mathcal{T}}, \hat{\mathcal{D}}] \bigr)
    \!=\! \{m, (R_i, \uts{D}_i, \uts{P}_i)_{i=1}^N) \}
     \text{,}  
\end{equation}
where $m \!\in\! \R$ is the predicted global metric scaling factor, and for each view $i$,
$R_i \!\in\! \R^{3 \times H \times W}$ are the predicted local ray directions,
$\uts{D}_i \!\in\! \R^{1 \times H \times W}$ are the ray depths in a up-to-scale space (indicated by the tilde),
and $\uts{P}_i \!\in\! \R^{4 \times 4}$ is the pose of image $\hat{I}_i$ in the frame of image $\hat{I}_1$, represented as quaternion $Q_i$
and up-to-scale translation $\uts{T}_i \!\in\! \R^3$.
We can further use this factored output to get the up-to-scale local pointmaps (3D points corresponding to each pixel) as $\uts{L}_i \!=\! R_i \!\cdot\! \uts{D}_i \!\in\! \R^{3 \times H \times W}$.
Then, leveraging the rotation matrix ${O}_i$ 
(obtained from $Q_i$) and up-to-scale translation, we can compute the $N$-view up-to-scale pointmaps in world frame as $\uts{X}_i \!=\! {O}_i \!\cdot\! \uts{L}_i \!+\! \uts{T}_i$.
The final metric 3D reconstruction for the $N$ input views (in the frame of image $I_1$) is given by
$\metric{X}_i \!=\! m \!\cdot\! \uts{X}_i$ for $i \!\in\! [1, N]$.

\subsection{Encoding Images \& Geometric Inputs}

Given $N$ visual inputs and optional dense geometric inputs, we first encode them into a common latent space.
For images, we use DINOv2 (Apache 2.0) \cite{OquabDMVSKFHMEABGHHLMRSSXJMLJB2024}.
Among a wide variety of pre-trained options, such as CroCov2 \cite{WeinzLLCABCACR2023}, DUSt3R's image encoder \cite{WangLCCR2024}, RADIO \cite{RanziHKM2024, HeinrRHLKTCM2025}, and random-init linear patchification, we find DINOv2 to be optimal in terms of downstream performance, convergence speed, and generalization (especially when fine-tuned with a small learning rate). 
We use the $24$th layer normalized patch features from DINOv2 ViT-G, $F_\text{I} \!\in\! \R^{1536 \times H/14 \times W/14}$.

\coolname can also encode other geometric quantities.
Before feeding these geometric quantities to our network, we factorize them to enable training and inference across both metric and up-to-scale quantities.
To support use cases where only rotation or translation might be individually present (for e.g., IMU \& GPS priors) and to deal with the entanglement of translation with scale, we encode rotation and translation separately.
Furthermore, since we don't assume depth \& pose to always be provided together as input, we decouple their normalization (note that this is separate from the training objective where we normalize predicted depth \& pose together since we want multi-view consistency).

In particular, when provided, the ray depths are first decoupled into average per-view depth $\hat{z}_{di} \!\in\! \R^+$ and normalized ray depths $\hat{D}_i / \hat{z}_{di}$.
Furthermore, when translations $\hat{\mathcal{T}}$ are provided, \coolname computes the pose scale as the average distance to the world frame, %
$\hat{z}_\text{p} \!=\! \frac{1}{\abs{S_t}} \sum_{i\in{S_t}} \lVert \hat{T}_i \rVert$.
This pose scale is used as the same input for all frames with input translation and is also used to get the normalized translations $\hat{T}_i / \hat{z}_\text{p}$.
Since we are interested in effectively exploiting the metric scale information from geometric inputs, \coolname{} only uses the pose scale and depth scales when the poses and depths provided for specific frames are metric.
Furthermore, the metric scale values can be large and drastically vary across scene sizes, hence, we log-transform scales before encoding them.

We encode ray directions and normalized ray depths using a shallow convolutional encoder \cite{MouWXWZQSQ2024}, where the spatial resizing only happens once with a pixel unshuffle of size 14.
This projects the dense geometric inputs into the same spatial and latent dimension as the DINOv2 features, \ie, $F_\text{R}, F_\text{D} \!\in\! \R^{1536 \times H/14 \times W/14}$.
For the global non-pixel quantities, i.e, rotations (represented as unit quaternions), translation directions, depth and pose scales, we use a 4-layer MLP with GeLU activations to project the quantities to features $F_\text{Q}, F_\text{T}, F_{\hat{z}_d}, F_{\hat{z}_\text{p}} \!\in\! \R^{1536}$.
Once all input quantities are encoded, they are passed through layer normalization, summed together, and followed by another layer normalization to obtain the final per-view encodings for each input view.
These are then flattened into tokens $F_\text{E} \!\in\! \R^{1536 \times ({HW}/{256})}$.

We append a single learnable scale token to the set of $N$ view patch tokens and input the tokens into a multi-view transformer to allow information across multiple views to attend to each other and propagate. 
We use a 16-layer alternating-attention transformer \cite{WangCKVRN2025} with 24~heads of multi-headed attention, a latent dimension of 1536 and an MLP ratio of 4, initialized using the last 16 layers of DINOv2 ViT-G \cite{oquab2023dinov2, lin2025depth}.
To distinguish the reference view (i.e., the first one), we add a constant reference view embedding to the set of patch tokens corresponding to view $I_{1}$.
For simplicity, we do not use Rotary Positional Embedding (RoPE) \cite{SuLPMWL2024}.
We find that the patch-level positional encoding from DINOv2 suffices, and RoPE leads to unnecessary biases, given that it was originally applied in every attention layer.

\subsection{Factored Scene Representation Prediction}

Once the multi-view transformer fuses information across different views and outputs the $N$-view patch tokens and scale token, \coolname{} further decodes these tokens into factored quantities representing the metric 3D geometry.
In particular,
we use a DPT head \cite{RanftBK2021} to decode the $N$-view patch tokens into $N$ dense per-view outputs, \ie, ray directions $R_i$ (normalized to unit length), up-to-scale ray depths $\uts{D}_i$,
masks $M_i$ representing non-ambiguous classes for depth,
and world-frame pointmap confidence maps $C_i$.
Furthermore, we input the $N$-view patch tokens into an average pooling-based convolutional pose head \cite{ChenCPB2024} to predict the unit quaternions $Q_i$ and up-to-scale translations $\uts{T}_i$.
Finally, the scale token is passed through a 2-layer MLP with ReLU activations to predict the metric scaling factor.
Since the metric scale of a scene can vary vastly, we exponentially scale the prediction to obtain the metric scaling factor $m$.
As shown in \cref{tab:ablation_rep}, we find that this decoupling of scale prediction is critical to achieving universal metric feed-forward inference.
Finally, as mentioned earlier, %
these factored predictions can be used together to obtain the metric 3D reconstruction.

\begin{figure*}[!t]
    \centering
    \includegraphics[trim={0cm 0.15cm 0cm 0cm},clip,width=0.94\linewidth]{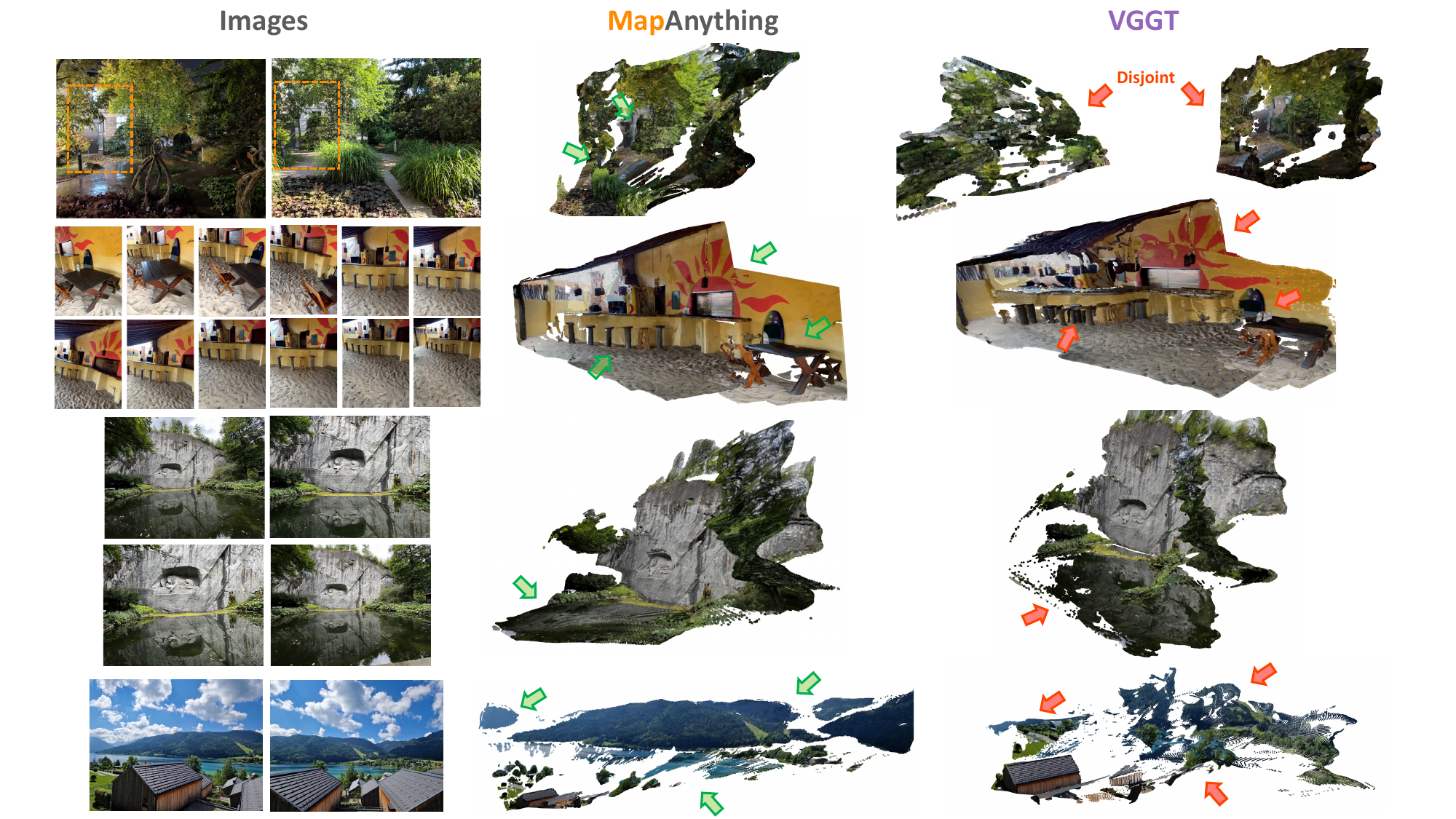}
    \caption{\textbf{Qualitative comparison of \coolname to VGGT \cite{WangCKVRN2025} using only in-the-wild images as input.}
    For a fair comparison, we apply the same normal-based edge mask post-processing and our sky mask to both methods. %
    \coolname more effectively deals with large disparity changes, seasonal shifts, textureless surfaces, water bodies and large scenes.
    }
    \label{fig:qual_comparison}
\end{figure*}

\subsection{Training Universal Metric 3D Reconstruction}
\label{sec:training}

We train \coolname end-to-end using multiple losses depending on the available supervision.
Since ray directions $R_i$ and pose quaternions $Q_i$ do not depend on scene scale, their losses are:
$\mathcal{L}_\text{rays} \!=\! \sum_{i=1}^N \| \hat{R}_i \!-\! R_i \|$ and
$\mathcal{L}_\text{rot} \!=\! \sum_{i=1}^N \min(\| \hat{Q}_i \!-\! Q_i \|, \| {-}\hat{Q}_i \!-\! Q_i \|)$.
This accounts for the two-to-one mapping of unit quaternions, and the regression loss is similar to a geodesic angular distance.

For the predicted up-to-scale ray depths $\uts{D}_i$, pose translations $\uts{T}_i$, local pointmaps $\uts{L}_i$ and world frame pointmaps $\uts{X}_i$, we follow DUSt3R \cite{WangLCCR2024} and use the ground-truth validity masks $V_i$ to compute the scaling factors for the ground truth $\hat{z} \!=\! \| (\hat{X}_i[V_i])_{i=1}^{N} \| / \sum_{i=1}^N \!V_i$ and the up-to-scale predictions $\uts{z} \!=\! \| (\uts{X}_i[V_i])_{i=1}^{N} \| / \sum_{i=1}^N \!V_i$. 
Likewise, to ensure that gradients from the scale loss do not influence the geometry, we use the predicted metric scaling factor $m$ and detached up-to-scale norm scaling factor $\uts{z}$ to compute the metric norm scaling factor $\metric{z} \!=\! m \!\cdot\! \mathrm{sg}(\uts{z})$, where $\mathrm{sg}$ indicates stop-grad.

Given these scaling factors, we compute the scale-invariant translation loss as $\mathcal{L}_\text{translation} \!=\! \sum_{i=1}^N \| \hat{T}_i/\hat{z} \!-\! \uts{T}_i/\uts{z} \|$.
We find that it is critical to apply losses in log-space for ray depths, pointmaps and the metric scale factor.
Specifically, we use $f_\mathrm{log} \colon \mathbf{x} \to (\mathbf{x}/\|\mathbf{x}\|) \!\cdot\! \log(1 \!+\! \|\mathbf{x}\|)$.
Thus, the loss for the ray depths is $\mathcal{L}_\text{depth} \!=\! \sum_{i=1}^N \| f_\mathrm{log}(\hat{D}_i/\hat{z}) \!-\! f_\mathrm{log}(\uts{D}_i/\uts{z}) \|$.
Likewise, the loss for the local pointmaps is $\mathcal{L}_\text{lpm} \!=\! \sum_{i=1}^N \| f_\mathrm{log}(\hat{L}_i/\hat{z}) \!-\! f_\mathrm{log}(\uts{L}_i/\uts{z}) \|$.
We exclude the top 5\% of per-pixel loss values to ignore imperfections and potential outliers in the training data.
Similar to DUSt3R, we add $\mathcal{L}_\text{pointmap} \!=\! \sum_{i=1}^N (C_i \| f_\mathrm{log}(\hat{X}_i/\hat{z}) \!-\! f_\mathrm{log}(\uts{X}_i/\uts{z}) \| \!-\! \alpha \log(C_i))$ as a confidence-weighted pointmap loss.
Lastly, the factored metric scale loss is given by $\mathcal{L}_\text{scale} \!=\! \| f_\mathrm{log}(\hat{z}) \!-\! f_\mathrm{log}(\metric{z}) \|$.

To capture fine details, %
we also employ a normal loss $\mathcal{L}_\text{normal}$ \cite{WangXDXDTY2025} on the local pointmaps, and a multi-scale gradient matching loss $\mathcal{L}_\text{GM}$ \cite{RanftLHSK2022, YangKHZXFZ2024} on the log of the $z$-depth in the local pointmaps.
Since the geometry from real datasets can be coarse and noisy, we apply the $\mathcal{L}_\text{normal}$ and $\mathcal{L}_\text{GM}$ losses only to synthetic datasets.
For the predicted non-ambiguous class masks, we use a binary cross entropy loss ($\mathcal{L}_\text{mask}$).

\noindent Overall, we use the following total loss: 
\begin{equation}
\begin{split}
    \label{eq:training_loss}
    \mathcal{L}
    =
    &10 \cdot \mathcal{L}_\text{pointmap}
    + \mathcal{L}_\text{rays}
    + \mathcal{L}_\text{rot}
    + \mathcal{L}_\text{translation}
    + \mathcal{L}_\text{depth}
    \\
    &+ \mathcal{L}_\text{lpm}
    + \mathcal{L}_\text{scale}
    + \mathcal{L}_\text{normal}
    + \mathcal{L}_\text{GM}
    + 0.1 \cdot \mathcal{L}_\text{mask}
\end{split}
\end{equation}
For the factored predictions, we find that up-weighting the global pointmap loss and down-weighting the mask loss is beneficial.
For all the regression losses, we use an adaptive robust loss \cite{Barro2019} (with parameters $c=0.05$ and $\alpha=0.5$) to help with robustness to outliers.

\paragraph{Training for Image \& Geometric Inputs:}
\label{para:aug_training}
To enable one-shot training of a universal model that supports various input configurations, we provide additional geometric inputs to the model with varying selection probabilities during training.
Specifically, we use an overall geometric input probability of 0.9, where each individual factorization, \ie, ray directions, ray depth, and pose, has an input probability of 0.5 each.
Whenever depth is selected as input, there is an equal probability of providing dense depth or 90\% randomly sparsified depth.
For robustness and flexibility in terms of which views have geometric information available as input, we use a per-view input probability of 0.95 and do not provide metric scale factors as input for metric-scale ground-truth datasets with a probability of 0.05.
We provide further details regarding the training setup in the supplement.

\input{tables/datasets}

\paragraph{Datasets:}
We train \coolname on 13 high-quality datasets (see \cref{tab:datasets}) with diversity across indoor, outdoor, and in-the-wild scenes.
For ScanNet++ v2 and TartanAirV2-WB, we split the scenes into a training, validation, and a held-out test set, while other datasets are split into training and validation.
While MPSD is originally a monocular metric depth dataset, we acquire the pose and camera information to enable a real-world multi-view metric scale dataset with $\sim$72K scenes.
We open-sourced this MPSD metadata to enable future research.
We release two pretrained models: one licensed under Apache 2.0 trained on six datasets, and one licensed under CC BY-NC 4.0 trained on an additional seven datasets (see \cref{tab:datasets}).
We provide comparisons between both variants in the supplementary.

\paragraph{Multi-View Sampling:}
For each dataset, we exhaustively precompute the pairwise covisibility of all images in a scene using a reprojection error check based on ground-truth depth and pose.
During training, we use this precomputed covisibility with a selected covisibility threshold of 25\% to perform random walk sampling.
This enables us to sample random single-connected component graphs of covisible views that have varying coverage and mutual information.

%% file: tables/datasets.tex
\begin{table}[b]
\caption{\label{tab:datasets}%
    \textbf{Datasets used for training and testing \coolname.}
}
\scriptsize
\centering
\newcommand{\fn}[1]{\textsuperscript{#1}}
\setlength\tabcolsep{5pt}
\begin{tabular}{llrc}
\toprule
\textbf{Dataset}                                          & \textbf{License}      & \textbf{\# Scenes} & \textbf{Metric} \\\midrule
BlendedMVS~\cite{YaoLLZRZFQ2020}                          & CC BY 4.0             &        493         & \xmark    \\
Mapillary Planet-Scale Depth~\cite{LopezGHBKK2020}                        & CC BY-NC-SA\fn{1}     &     71,428         & \tickmark \\
ScanNet++ v2~\cite{YeshwLND2023}                          & Non-commercial\fn{1}  &        926         & \tickmark \\
Spring~\cite{MehlSJNB2023}                                & CC BY 4.0             &         37         & \tickmark \\
TartanAirV2-WB~\cite{WangZWHQWHKS2020, ZhangKLJCQKJHRSW2025}     & CC BY 4.0             &         49         & \tickmark \\
UnrealStereo4K~\cite{TosiLSG2021}                         & MIT                   &          9         & \tickmark \\[.5em]
\multicolumn{4}{l}{\textbf{Additionally used for our CC BY-NC model:}} \\
Aria Synthetic Environments~\cite{AvetiXHYAPZFHOEMNB2024} & Non-commercial        &    103,890         & \tickmark \\
DL3DV-10K~\cite{LingSTZXWYGYLLSAMKKHZBB2024}              & CC BY-NC 4.0          &     10,109         & \xmark    \\
Dynamic Replica~\cite{KaraeRGNVR2023}                     & Non-commercial        &        523         & \tickmark \\
MegaDepth~\cite{LiS2018}                                  & CC BY 4.0\fn{2}       &        269         & \xmark    \\
MVS-Synth~\cite{HuangMKAH2018}                            & Non-commercial        &        120         & \tickmark \\
ParallelDomain-4D~\cite{VanHWOSLTDZV2024}                 & Non-commercial        &      1,528         & \tickmark \\
SAIL-VOS 3D~\cite{HuWYS2021}                              & Non-commercial        &        171         & \tickmark \\

\arrayrulecolor{gray}\midrule

\multicolumn{4}{l}{\textbf{Unique held-out scenes for dense up-to-N-view benchmarking:}} \\
ETH3D~\cite{SchoepSGSSPG2017}                             & CC BY-NC-SA 4.0       &         13         & \tickmark \\
ScanNet++ v2~\cite{YeshwLND2023}                          & Non-commercial\fn{1}  &         30         & \tickmark \\
TartanAirV2-WB~\cite{WangZWHQWHKS2020, ZhangKLJCQKJHRSW2025}     & CC BY 4.0             &          5         & \tickmark \\
\arrayrulecolor{black}\bottomrule
\end{tabular}\\[.2em]
\scriptsize{\fn{1} We obtained approval from the dataset owners that allows training and model release under a permissive license.
\quad
\fn{2} Crowd-sourced images with varying licenses.}
\end{table}

%% file: text/04_benchmarking.tex
\section{Benchmarking \& Results}
\label{sec:benchmarking}

In this section, we benchmark \coolname across a wide suite of 3D vision tasks.
For each task, we compare against expert baselines specifically designed or trained for the task.
We perform all experiments with a constant seed.

\paragraph{Multi-View Dense Reconstruction:}

\begin{figure*}[!t]
    \centering
    \includegraphics[trim={0cm 0cm 0cm 0.9cm},clip,width=\linewidth]{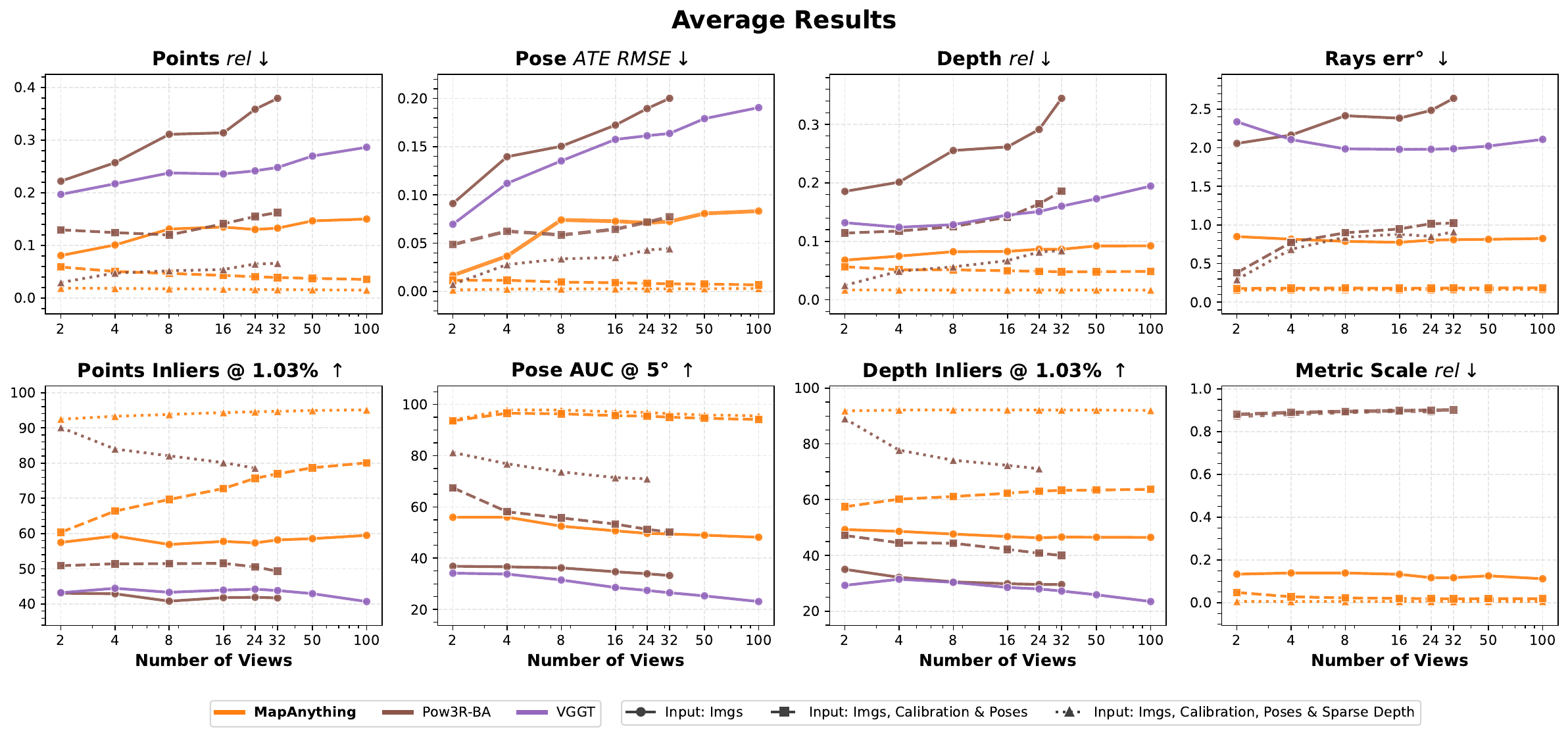}
    \caption{\label{fig:comparison_graphs}%
    \textbf{\coolname shows state-of-the-art dense multi-view reconstruction for input views ranging from 2 to 100 and under different input configurations.}
    We report the absolute relative error ($\absrel$), the inlier ratio at a relative threshold of $1.03\%$ ($\threshI$), the average aligned trajectory error ($ATE~RMSE$), the area under the curve at an error threshold of 5° ($AUC@5$), and the average angular error ($err$) in degrees (°), averaged over ETH3D, ScanNet++ v2 \& TAv2.
    We do not report performance for baselines when the inference runs out of GPU memory.
    We provide results for individual datasets \& the exhaustive input configurations of \coolname in the supplement.
    }
\end{figure*}

We benchmark the performance of pointmaps, pose, depth \& ray direction estimation on an undistorted version of ETH3D \cite{SchoepSGSSPG2017}, ScanNet++ v2 \cite{YeshwLND2023}, and TartanAirV2-WB \cite{WangZWHQWHKS2020, ZhangKLJCQKJHRSW2025}, where, for each test scene, we randomly sample up to $N$ views that form a single connected component graph based on the pre-computed pairwise covisibility of all images in the scene (this prevents disjoint sets of images as input).
\cref{fig:comparison_graphs} shows that \coolname provides state-of-the-art dense multi-view reconstruction performance over other baselines using only image input, including VGGT \cite{WangCKVRN2025}.
Beyond the performance using only images as input, we show that \coolname can leverage additional auxiliary geometric inputs for feed-forward inference to further increase reconstruction performance by a significant factor.
Furthermore, we find that \coolname is better than the bundle adjustment (BA) variant of the two-view baseline, Pow3R \cite{JangWLAR2025}, which is also designed to leverage scene priors.
We also find that reconstruction outputs from \coolname (using only images as input) display high fidelity, as shown in \cref{fig:qual_comparison}. 

\paragraph{Two-View Dense Reconstruction:}

We benchmark sparse-view reconstruction and image-matching performance against state-of-the-art feed-forward baselines in \cref{tab:two_view_dense_recon}.
\coolname achieves state-of-the-art performance using only images as input.
With additional input modalities, \coolname significantly outperforms both image-only baselines and Pow3R \cite{JangWLAR2025}, the only other two-view feed-forward method that uses scene or camera priors.

\input{tables/two_view_dense_recon}

\paragraph{Single-View Calibration:}

We benchmark the single-view calibration performance of \coolname and other expert calibration baselines on randomly sampled frames from the test scenes of undistorted ETH3D \cite{SchoepSGSSPG2017}, ScanNet++ v2 \cite{YeshwLND2023}, and TartanAirV2 \cite{WangZWHQWHKS2020}.
To test non-centered principal points, we randomly crop frames with aspect ratios from 3:1 to 1:2.
Despite not being trained specifically on single images, \cref{tab:single_view_calib} shows that \coolname achieves state-of-the-art performance for perspective calibration.
This demonstrates \coolname's effectiveness in modeling generic central camera systems
and its potential to generalize to wide-angle models like fisheye with appropriate training.

\input{tables/sv_calib}

\input{tables/robust_mvd}

\paragraph{Monocular \& Multi-View Depth Estimation:}
\label{subsec:depth_estimation}

In \cref{tab:robust-multi-view-depth}, we benchmark \coolname against specialized models for single-view and multi-view depth estimation across various inputs.
In the RMVD benchmark, note that we don't use ETH3D due to the distortion issue mentioned in MVSA~\cite{IzquiSFGTCABW2025} and DTU \& Tanks and Temples since they are not metric.
Although not trained specifically for single-view metric depth, \coolname achieves state-of-the-art or comparable performance.
For multi-view metric depth estimation using images only, \coolname outperforms MASt3R-BA \cite{LeroyCR2024} \& MUSt3R \cite{CabonSACCRL2025}.
With auxiliary inputs like camera calibration and poses, \coolname's performance improves and it delivers competitive results compared to task-specific specialized models.
In comparison to baselines such as MoGe-2 \cite{WangXDDXLSTY2025} and MVSA \cite{IzquiSFGTCABW2025}, we find the metric scale estimation on ScanNet to be sub-optimal and believe this is likely due to lower benchmark dataset quality \cite{WangXDXDTY2025, IzquiSFGTCABW2025}.
As indicated in \cref{tab:robust-multi-view-depth}, we observe strong depth estimation performance on ScanNet when using median scale alignment.

\paragraph{Insights into enabling \coolname:}
\label{para:insights}

As shown in \cref{tab:ablation_rep}, the factored representation of the scene as a multi-view set of rays, depth \& pose (RDP) along with the metric scale is a key enabler for strong reconstruction performance while using images and optionally additional geometric inputs.
In \cref{tab:ablation_uni_training}, we find that our input probability-based training is efficient in training one universal model for various tasks and input configurations, where the performance of the universally trained model is equivalent to various bespoke models trained for specific input configurations.

\input{tables/ablations_rep_uni_training}  %

%% file: tables/two_view_dense_recon.tex
\begin{table}[t]
\caption{\label{tab:two_view_dense_recon}%
  \textbf{\coolname showcases state-of-the-art two-view reconstruction under different input configurations.} We report the absolute relative error ($\absrel$), the inlier ratio at a relative threshold of 1.03\% ($\threshI$), the average aligned trajectory error (ATE), the area under the curve at an error threshold of 5° (AUC), and the average angular error ($err$) in degrees (°). Best results are indicated in \textbf{bold}.
}
\centering
\scriptsize
\setlength\tabcolsep{2pt}
\resizebox{\columnwidth}{!}{%
\begin{tabular}{l@{\hspace{8pt}}c@{\hspace{8pt}}c@{\hspace{6pt}}c@{\hspace{8pt}}cc@{\hspace{8pt}}c@{\hspace{6pt}}c@{\hspace{8pt}}c}
\toprule
& \multicolumn{8}{c}{\textbf{Average across ETH3D, SN++v2 \& TAV2}} \\
\cmidrule(lr){2-9}
& \multicolumn{1}{c@{\hspace{8pt}}}{\textbf{Scale}}
& \multicolumn{2}{c@{\hspace{8pt}}}{\textbf{Points}}
& \multicolumn{2}{c@{\hspace{8pt}}}{\textbf{Pose}}
& \multicolumn{2}{c@{\hspace{8pt}}}{\textbf{Depth}}
& \multicolumn{1}{c}{\textbf{Rays}} \\
\textbf{Methods}
& $\absrel\downarrow$
& $\absrel\downarrow$
& $\threshI\uparrow$
& \scriptsize ATE $\downarrow$
& \scriptsize AUC $\uparrow$
& $\absrel\downarrow$
& $\threshI\uparrow$
& err° $\downarrow$ \\
\midrule
\multicolumn{9}{l}{\textbf{a) Input: Images}} \\
DUSt3R~\cite{WangLCCR2024}     & ---   & 0.21    & 43.9    & 0.08   & 35.5    & 0.17   & 32.6    & 2.55    \\
MASt3R~\cite{LeroyCR2024}     & 0.38   & 0.25    & 30.2    & 0.07   & 37.3    & 0.19   & 24.8    & 7.03    \\
Pow3R~\cite{JangWLAR2025}     & ---   & 0.22    & 43.1    & 0.09   & 36.9    & 0.19   & 35.0    & 2.06    \\
VGGT~\cite{WangCKVRN2025}     & ---   & 0.20    & 43.2    & 0.07   & 34.2    & 0.13   & 29.3    & 2.34    \\
\textbf{{\coolname}}     & \textbf{0.13}   & \textbf{0.08}    & \textbf{57.5}    & \textbf{0.02}   & \textbf{56.0}    & \textbf{0.07}   & \textbf{49.3}    & \textbf{0.85}    \\

\arrayrulecolor{gray}\midrule
\multicolumn{9}{l}{\textbf{b) Input: Images \& Intrinsics}} \\
Pow3R~\cite{JangWLAR2025}     & ---   & 0.20    & 46.0    & 0.08   & 51.3    & 0.15   & 43.2    & 0.40    \\
\textbf{{\coolname}}     & \textbf{0.13}   & \textbf{0.07}    & \textbf{59.3}    & \textbf{0.01}   & \textbf{64.7}    & \textbf{0.06}   & \textbf{55.1}    & \textbf{0.19}    \\

\midrule
\multicolumn{9}{l}{\textbf{c) Input: Images, Intrinsics \& Poses}} \\
Pow3R~\cite{JangWLAR2025}     & ---   & 0.13    & 50.9    & 0.05   & 67.5    & 0.11   & 47.2    & 0.38    \\
\textbf{{\coolname}}     & \textbf{0.05}   & \textbf{0.06}    & \textbf{60.4}    & \textbf{0.01}   & \textbf{93.6}    & \textbf{0.06}   & \textbf{57.5}    & \textbf{0.18}    \\

\midrule
\multicolumn{9}{l}{\textbf{d) Input: Images, Intrinsics \& Depth}} \\
Pow3R~\cite{JangWLAR2025}     & ---   & 0.13    & \textbf{77.9}    & 0.04   & 66.5    & 0.07   & \textbf{77.3}    & 0.29    \\
\textbf{{\coolname}}     & \textbf{0.02}   & \textbf{0.04}    & 77.8    & \textbf{0.01}   & \textbf{73.1}    & \textbf{0.03}   & 76.6    & \textbf{0.18}    \\

\midrule
\multicolumn{9}{l}{\textbf{e) Input: Images, Intrinsics, Poses \& Depth}} \\
Pow3R~\cite{JangWLAR2025}     & ---   & 0.03    & \textbf{90.1}    & 0.01   & 81.3    & 0.02   & \textbf{89.0}    & 0.29    \\
\textbf{{\coolname}}     & \textbf{0.01}   & \textbf{0.02}    & 82.0    & \textbf{0.00}   & \textbf{94.8}    & \textbf{0.02}   & 81.5    & \textbf{0.16}    \\

\arrayrulecolor{black}\bottomrule
\end{tabular}%
}
\end{table}

%% file: tables/sv_calib.tex
\begin{table}[t]
\caption{\label{tab:single_view_calib}%
    \textbf{\coolname shows state-of-the-art single-image calibration.} Note that \coolname has not been trained specifically for single-image inputs. We report the average angular error ($err$) in degrees (°). Best results are indicated in \textbf{bold}.
}
\centering
\scriptsize
\begin{tabular}{lcccc}
\toprule

    \textbf{Methods}
    & \multicolumn{1}{c}{\textbf{Avg.}}
    & \multicolumn{1}{c}{\textbf{ETH3D}}
    & \multicolumn{1}{c}{\textbf{SN++v2}}
    & \multicolumn{1}{c}{\textbf{TAV2}}
    \\

    \midrule

    VGGT~\cite{WangCKVRN2025}
    & 4.00
    & 2.83
    & 5.21
    & 3.95
    \\

    MoGe-2~\cite{WangXDDXLSTY2025}
    & 1.95
    & 1.89
    & 1.56
    & 2.40
    \\

    AnyCalib~\cite{TiradC2025}
    & 2.01
    & 1.52
    & 2.41
    & 2.10
    \\

    \textbf{\coolname}
    & \textbf{1.06}
    & \textbf{1.33}
    & \textbf{0.39}
    & \textbf{1.47}
    \\

\bottomrule
\end{tabular}
\end{table}

%% file: tables/robust_mvd.tex
\begin{table}[t]
    \caption{\label{tab:robust-multi-view-depth}%
        \textbf{\coolname shows versatile metric depth estimation under different input configurations on the Robust-MVD Benchmark \cite{SchroeBAB2022}}.
        Note that \coolname has not been trained for single-image inputs.
        We report the absolute relative error ($\absrel$) and the inlier ratio at a relative threshold of 1.03\% ($\threshI$).
        The best result for each group is in \bestresult{bold}; \textcolor{lightgray}{gray text} indicates results where the evaluation dataset is in the training distribution~\cite{IzquiSFGTCABW2025}.
    }
    \scriptsize
    \centering
        \input{tables/robust_mvd_numbers_full}
\end{table}

%% file: tables/robust_mvd_numbers_full.tex
\def\pad{\phantom{0}}
\begin{tabular}{l
@{\hspace{6pt}}c
@{\hspace{6pt}}c
c c
c c
}

\toprule
    &
    &
    & \multicolumn{2}{c}{\textbf{KITTI}}
    & \multicolumn{2}{c}{\textbf{ScanNet}}
    \\

    \textbf{Approach}
    & \textbf{K}
    & \textbf{Poses}
    & $\absrel\downarrow$ & $\threshI\uparrow$
    & $\absrel\downarrow$ & $\threshI\uparrow$
    \\

    \midrule

    \multicolumn{7}{l}{\textbf{a) Single-View Metric}}
    \\[0.4em]

        MoGe-2~\cite{WangXDDXLSTY2025}
        & \xmark
        & \xmark
        & 14.21
        & \pad 6.8
        & \pad \textbf{10.57}
        & \textbf{19.8}
        \\

        \textbf{\coolname}
        & \xmark
        & \xmark
        & \pad\textbf{9.69}
        & \textbf{17.9}
        & \pad 27.77
        & \pad 2.9
        \\

    \arrayrulecolor{lightgray}\midrule

        Depth Pro~\cite{BochkDGSZRK2025}
        & \tickmark
        & \xmark
        & 13.60
        & 14.3
        & \pad\pad\textcolor{lightgray}{9.20}
        & \textcolor{lightgray}{19.7}
        \\

        UniDepthV2~\cite{PicciSYSLAV2026}
        & \tickmark
        & \xmark
        & 13.70
        & \pad 4.8
        & \pad\pad\textcolor{lightgray}{3.20}
        & \textcolor{lightgray}{61.3}
        \\

        Metric3DV2~\cite{HuYZCLCWYSS2024}
        & \tickmark
        & \xmark
        & \pad8.70
        & 13.2
        & \pad\pad\textcolor{lightgray}{6.20}
        & \textcolor{lightgray}{19.3}
        \\

        \textbf{\coolname}
        & \tickmark
        & \xmark
        & \pad\textbf{8.48}
        & \textbf{27.7}
        & \pad\textbf{31.12}
        & \pad\textbf{3.0}
        \\

    \arrayrulecolor{gray}\midrule

    \multicolumn{7}{l}{\textbf{b) Multi-View Metric}}
    \\[0.4em]

        MAST3R~\cite{LeroyCR2024}
        & \xmark
        & \xmark
        & 61.40 %
        & \pad 0.4
        & \pad\textcolor{lightgray}{12.80} %
        & \textcolor{lightgray}{19.4}
        \\

        MUSt3R~\cite{CabonSACCRL2025}
        & \xmark
        & \xmark
        & 19.76
        & \pad 7.3
        & \pad\pad\textcolor{lightgray}{7.66}
        & \textcolor{lightgray}{35.7}
        \\ %

        \textbf{\coolname}
        & \xmark
        & \xmark
        & \pad\textbf{5.45}
        & \textbf{45.7}
        & \pad\textbf{22.23}
        & \textbf{10.6}
        \\

    \arrayrulecolor{lightgray}\midrule

        \textbf{\coolname}
        & \tickmark
        & \xmark
        & \pad\textbf{8.45}
        & \textbf{27.5}
        & \pad\textbf{24.94}
        & \pad\textbf{8.2}
        \\

    \arrayrulecolor{lightgray}\midrule

        Fast-MVSNet~\cite{YuG2020}
        & \tickmark
        & \tickmark
        & 12.10
        & 37.4
        & 287.10
        & \pad 9.4
        \\ %

        Robust MVDB~\cite{SchroeBAB2022}
        & \tickmark
        & \tickmark
        & \pad 7.10
        & 41.9
        & \pad\pad 7.40
        & 38.4
        \\ %
    
        MAST3R Tri.~\cite{IzquiSFGTCABW2025}
        & \tickmark
        & \tickmark
        & \pad 3.40
        & 66.6
        & \pad\pad\textcolor{lightgray}{4.50}
        & \textcolor{lightgray}{63.0}
        \\ %
    
        MVSA~\cite{IzquiSFGTCABW2025}%
        & \tickmark
        & \tickmark
        & \pad\bestresult{3.20}
        & \bestresult{68.8}
        & \pad \pad\bestresult{3.70}
        & \bestresult{62.9}
        \\

        \textbf{\coolname}
        & \tickmark
        & \tickmark
        & \pad 4.63
        & 51.6
        & \pad \pad 5.58
        & 48.1
        \\ %

    \midrule

    \multicolumn{7}{l}{\textbf{c) Single-View w/ Alignment}}
    \\[0.4em]

        MoGe~\cite{WangXDXDTY2025}
        & \xmark
        & \xmark
        & \pad 5.12
        & 46.2 %
        & \pad\pad\textbf{3.59}
        & \textbf{65.3} %
        \\

        MoGe-2~\cite{WangXDDXLSTY2025}
        & \xmark
        & \xmark
        & \pad\textbf{4.82}
        & \textbf{47.9} %
        & \pad\pad3.77
        & 63.1 %
        \\

        VGGT~\cite{WangCKVRN2025}
        & \xmark
        & \xmark
        & \pad7.50
        & 33.0 %
        & \pad\pad\textcolor{lightgray}{3.33}
        & \textcolor{lightgray}{70.8} %
        \\

        $\pi^3$~\cite{WangZZCZLCPSH2025}
        & \xmark
        & \xmark
        & \pad6.00
        & 40.1
        & \pad\pad\textcolor{lightgray}{2.90}
        & \textcolor{lightgray}{73.9}
        \\

        \textbf{\coolname}
        & \xmark
        & \xmark
        & \pad6.12
        & 42.2
        & \pad \pad 4.95
        & 55.6
        \\
        
    \arrayrulecolor{lightgray}\midrule

        Depth Pro~\cite{BochkDGSZRK2025}
        & \tickmark
        & \xmark
        & \pad 6.10
        & 39.6
        & \pad\pad\textcolor{lightgray}{4.30}
        & \textcolor{lightgray}{58.4}
        \\

        DAV2~\cite{YangKHZXFZ2024}
        & \tickmark
        & \xmark
        & \pad 6.60
        & 38.6
        & \pad\pad\textbf{4.00}
        & \textbf{58.6}
        \\ %

        Metric3DV2~\cite{HuYZCLCWYSS2024}
        & \tickmark
        & \xmark
        & \pad 5.10
        & 44.1
        & \pad\pad\textcolor{lightgray}{2.40}
        &  \textcolor{lightgray}{78.3}
        \\

        UniDepthV2~\cite{PicciSYSLAV2026}
        & \tickmark
        & \xmark
        & \pad\bestresult{4.00}
        & \bestresult{55.3}
        & \pad\pad\textcolor{lightgray}{2.10}
        & \textcolor{lightgray}{82.6}
        \\

        \textbf{\coolname}
        & \tickmark
        & \xmark
        & \pad6.15
        & 41.6
        & \pad\pad 4.77
        & 57.1
        \\

    \midrule

    \multicolumn{7}{l}{\textbf{d) Multi-View w/ Alignment}}
    \\[0.4em]

        MAST3R~\cite{LeroyCR2024}
        & \xmark
        & \xmark
        & \pad3.30
        & 67.7
        & \pad\pad\textcolor{lightgray}{4.30}
        & \textcolor{lightgray}{64.0}
        \\ %

        MUSt3R~\cite{CabonSACCRL2025}
        & \xmark
        & \xmark
        & \pad 4.47
        & 56.7
        & \pad\pad\textcolor{lightgray}{3.22}
        & \textcolor{lightgray}{69.2}
        \\ %

        VGGT~\cite{WangCKVRN2025}
        & \xmark
        & \xmark
        & \pad 4.60
        & 53.0 %
        & \pad\pad\textcolor{lightgray}{2.34}
        & \textcolor{lightgray}{80.6} %
        \\

        $\pi^3$~\cite{WangZZCZLCPSH2025}
        & \xmark
        & \xmark
        & \pad\textbf{3.09}
        & \textbf{69.5}
        & \pad\pad\textcolor{lightgray}{1.98}
        & \textcolor{lightgray}{83.6}
        \\

        \textbf{\coolname}
        & \xmark
        & \xmark
        & \pad 4.04
        & 60.3
        & \pad\pad\textbf{3.47}
        & \textbf{67.0}
        \\

    \arrayrulecolor{lightgray}\midrule

        DeMoN~\cite{UmmenZUMIDB2017}
        & \tickmark
        & \xmark
        & 15.50
        & 15.2
        & \pad 12.00
        & 21.0
        \\ %
    
        DeepV2D {\scriptsize{KITTI}}~\cite{TeedD2020}
        & \tickmark
        & \xmark
        & \pad\textcolor{lightgray}{3.10}
        & \textcolor{lightgray}{74.9}
        & \pad 23.70
        & 11.1
        \\ %
    
        DeepV2D {\scriptsize{ScanNet}}~\cite{TeedD2020}
        & \tickmark
        & \xmark
        & 10.00
        & 36.2
        & \pad\pad\textcolor{lightgray}{4.40}
        & \textcolor{lightgray}{54.8}
        \\ %

        \textbf{\coolname}
        & \tickmark
        & \xmark
        & \pad\textbf{3.97}
        & \textbf{61.2}
        & \pad\pad\textbf{3.34}
        & \textbf{68.5}
        \\ %

    \arrayrulecolor{black}\bottomrule
\end{tabular}

%% file: tables/ablations_rep_uni_training.tex
\begin{table}
  \centering
  \caption{\label{tab:ablations_rep_uni_training}%
    \textbf{Ablations providing insight into the key design choices.} %
    We report the absolute relative error ($\absrel$) and the inlier ratio at a relative threshold of 1.03\% ($\threshI$) at 50 views, averaged over ETH3D, ScanNet++ v2 \& TAv2.
    Best results are \textbf{bold}.
    \textbf{Insights:}
    (a) The factored representation of rays, depth \& pose (RDP) along with metric scale is key to achieving strong reconstruction performance under different input configurations.
    (b) \coolname trained universally for 12+ tasks in one go with equivalent compute to two bespoke models is superior in terms of performance to three bespoke models trained for distinct input configurations.
    This indicates that the multi-task training of \coolname is highly efficient.
  }
  \begin{subtable}[t]{0.49\linewidth}
    \centering
    \caption{\textbf{Scene Representation}}
    \input{tables/ablation_rep}

    \label{tab:ablation_rep}
  \end{subtable}%
  \hfill
  \begin{subtable}[t]{0.50\linewidth}
    \centering
    \caption{\textbf{Expert vs Universal Training}}
    \input{tables/ablation_uni_training}

    \label{tab:ablation_uni_training}
  \end{subtable}%
\end{table}

%% file: tables/ablation_rep.tex
\resizebox{\textwidth}{!}{
\begin{tabular}{lccc}
\toprule

    & \textbf{Metric Scale} & \multicolumn{2}{c}{\textbf{Pointmaps}}

    \\

    \textbf{Methods}
    & $\absrel\downarrow$
    & $\absrel\downarrow$ & $\threshI\uparrow$

    \\

    \midrule

    \multicolumn{4}{l}{\textbf{Input: Images Only}}
    \\

    Local PM + Pose
    & \textbf{0.14}
    & 0.32
    & 33.2
    \\

    RDP
    & 0.17
    & 0.33
    & 32.6
    \\

    LPMP \& Scale
    & 0.16
    & 0.30
    & 38.7
    \\

    RDP \& Scale (ours)
    & 0.16
    & \textbf{0.28}
    & \textbf{40.7}
    \\

    \arrayrulecolor{gray}\midrule

    \multicolumn{4}{l}{\textbf{Input: Images, Intrinsics \& Metric Poses}}
    \\

    Local PM + Pose
    & \textbf{0.04}
    & 0.08
    & 53.5
    \\

    RDP
    & 0.06
    & 0.09
    & 46.7
    \\

    LPMP \& Scale
    & 0.06
    & \textbf{0.07}
    & 55.9
    \\

    RDP \& Scale (ours)
    & 0.05
    & \textbf{0.07}
    & \textbf{57.8}
    \\

\arrayrulecolor{black}\bottomrule
\end{tabular}
}

%% file: tables/ablation_uni_training.tex
\resizebox{\textwidth}{!}{
\begin{tabular}{lccc}
\toprule

    & \textbf{Metric Scale} & \multicolumn{2}{c}{\textbf{Pointmaps}}

    \\

    \textbf{Methods}
    & $\absrel\downarrow$
    & $\absrel\downarrow$ & $\threshI\uparrow$

    \\

    \midrule

    \multicolumn{4}{l}{\textbf{Input: Images Only}}
    \\

    Expert Training
    & \textbf{0.16}
    & 0.29
    & 31.8
    \\

    Universal Training
    & \textbf{0.16}
    & \textbf{0.28}
    & \textbf{40.7}
    \\

    \arrayrulecolor{gray}\midrule

    \multicolumn{4}{l}{\textbf{Input: Images, Intrinsics \& Metric Poses}}
    \\

    Expert Training
    & \textbf{0.03}
    & \textbf{0.07}
    & 56.2
    \\

    Universal Training
    & 0.05
    & \textbf{0.07}
    & \textbf{57.8}
    \\

    \arrayrulecolor{gray}\midrule

    \multicolumn{4}{l}{\textbf{Input: Images \& Metric Depth}}
    \\

    Expert Training
    & \textbf{0.06}
    & \textbf{0.24}
    & 53.0
    \\

    Universal Training
    & \textbf{0.06}
    & 0.25
    & \textbf{54.0}
    \\

\arrayrulecolor{black}\bottomrule
\end{tabular}
}

%% file: text/05_conclusion.tex
\section{Limitations}
\label{sec:limitations}

While \coolname makes significant strides towards a universal multi-modal backbone for in-the-wild metric-scale 3D reconstruction, several limitations and future directions remain:
(a) \coolname does not explicitly account for noise or uncertainty in geometric inputs.
(b) Although this is not currently supported, the architecture can be easily extended to handle tasks where images are not available for all input views. For example, in novel view synthesis, the target views for rendering will only have cameras available as input.
(c) While the design of \coolname supports iterative inference, it remains to be explored how effective scaling of test-time compute would be for 3D reconstruction (this ties into effectively handling noise in the inputs).
(d) Multi-modal features are currently fused before being input; exploring more efficient ways to directly input different modalities to the transformer could be interesting.

Beyond multi-task capabilities, scalability is currently limited by the one-to-one mapping between input pixels and the output scene representation.
We believe that significant work remains in effectively representing scenes in memory and decoding them as required, especially for large scenes.
Our current scene parameterization does not capture dynamic motion or scene flow \cite{karhade2025any4d}, which are promising areas.

\section{Conclusion}
\label{sec:conclusion}

\coolname is the first universal transformer-based backbone that directly regresses metric 3D geometry and camera poses from flexible inputs -- including images, camera intrinsics, poses, depth maps, or partial reconstructions -- in a single pass. 
By using a factored representation of multi-view geometry (depth maps, ray maps, poses, and a global scale factor), \coolname unifies local estimates into a global metric frame.
\coolname handles multiple tasks like uncalibrated structure-from-motion, calibrated multi-view stereo, monocular depth estimation, camera localization, depth completion, and more without task-specific tuning.
Extensive experiments show that it surpasses or matches specialist models while enabling efficient joint training. 
Future extensions to dynamic scenes, uncertainty quantification, and scene understanding promise to further generalize \coolname's capabilities and robustness, paving the way toward a truly universal 3D reconstruction backbone.

%% file: supplementary.tex
\clearpage
\maketitlesupplementary
\appendix

\setcounter{section}{0}
\setcounter{equation}{0}
\setcounter{figure}{0}
\setcounter{table}{0}
\renewcommand{\thefigure}{S.\arabic{figure}}
\renewcommand{\thetable}{S.\arabic{table}}
\renewcommand{\theequation}{S.\arabic{equation}}

\input{text/06_acknowledgments}

\vspace{-0.8em}

\section{Implementation Details}

We use the AdamW \cite{KingmB2015} optimizer with a peak learning rate of $5\cdot10^{-6}$ for the pre-trained DINOv2 encoder \cite{OquabDMVSKFHMEABGHHLMRSSXJMLJB2024}, and $10^{-4}$ for everything else.
For the learning rate schedule, we employ a 10\% linear warmup to the peak and subsequently use a half-cycle cosine decay to a $100\times$ lower value.
For the optimizer, we also use a weight decay of 0.05, $\beta_1 \!=\! 0.9$, and $\beta_2 \!=\! 0.95$.
For every batch, the input images and dense geometric quantities are resized and cropped so that the maximum dimension is 518 pixels and aspect ratio is randomized from 3:1 to 1:2.
We use color jitter, Gaussian blur, and grayscale conversion as augmentations.
We further employ mixed precision training and gradient checkpointing for DINOv2 encoder to improve training efficiency and GPU memory utilization.
We also use gradient norm clipping with a threshold of 1 for additional training stability.
Lastly, we use a dynamic batching scheme where the batch size is changed based on the number of views in a batch.
We find that it is effective to train the model with a two-stage curriculum (420K steps):
(1) 6 days on 64 H200-140GB GPUs with an effective batch size varying from 768 to 1536 with the number of views varying from 4 to 2, respectively, and
(2) 4 days on 64 H200-140GB GPUs with a 10$\times$ lower peak LR and an effective batch size that varies from 128 to 1536 with views varying from 24 to 2, respectively.
The training setup for the ablations presented in \cref{para:insights} is the same as above where the only difference is the effective batch size (8$\times$ lower but leading to sufficient convergence).

\vspace{-0.7em}

\section{Additional Evaluation}

\begin{figure*}[t]
    \centering
    \begin{subfigure}{0.49\textwidth}
        \centering
        \includegraphics[width=\linewidth]{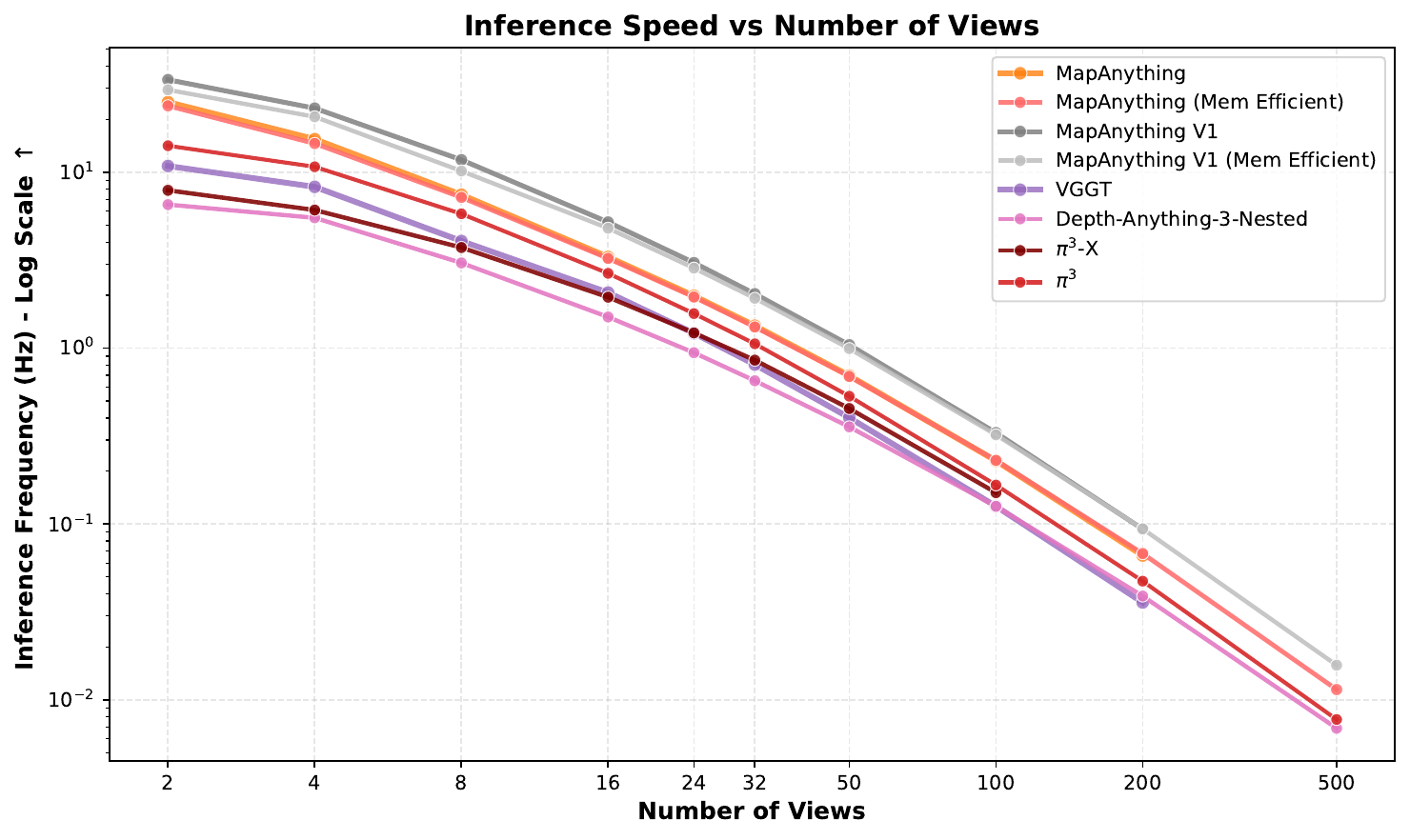}
    \end{subfigure}
    \hfill %
    \begin{subfigure}{0.49\textwidth}
        \centering
        \includegraphics[width=\linewidth]{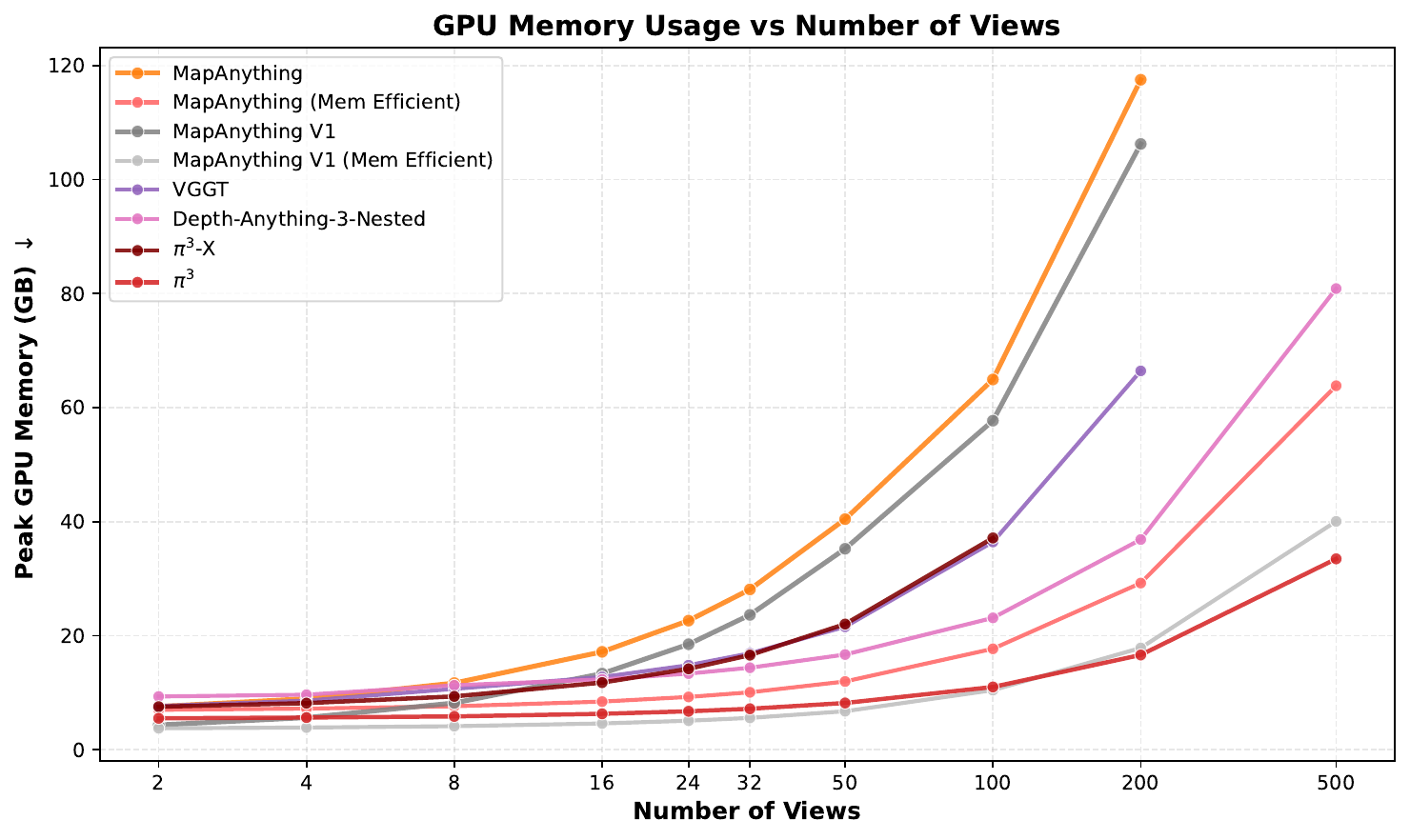}
    \end{subfigure}

    \caption{\label{fig:profiling}
    \textbf{\coolname shows the best speed and memory usage profile in comparison to latest concurrent state-of-the-art multi-view feed-forward reconstruction models.} 
    We profile all the models on a H200-140GB GPU using appropriate floating point precision autocast.
    Upon profiling the naive inference call of \coolname and comparing to concurrent methods, we notice a memory bottleneck in the dense per-pixel regression across views, since our implementation of the dense head is closer to the original DPT~\cite{RanftBK2021}.
    To alleviate this issue, i.e., in \coolname (Mem Efficient), we run the per-view decodings in a mini-batched loop, where in the above profiling the mini-batch size is 1.
    We find that this looping leads to negligible trade-off in speed while significantly reducing the memory usage.
    }
\end{figure*}

\paragraph{Speed \& Memory Profiling:} We present profiling results for \coolname and other concurrent models in \cref{fig:profiling}.

\vspace{-0.15em}

\paragraph{Flexibility of Input Configurations:}

We present a representative set of input configurations for \coolname in \cref{tab:supplementary_mapanything_input_variants}, where the performance of \coolname improves as more modalities are provided.
The universal training of \coolname with varying selection probabilities for geometric inputs as 6 factors (as described in \cref{para:aug_training}) enables support for 64 exhaustive input combinations.
While we primarily benchmark cases where the input modality is available for all views, \coolname can also support optional geometric inputs for a subset of the input views.

\vspace{-0.15em}

\paragraph{Comparison of \coolname Variants:}

\begin{figure*}[!t]
    \centering
    \includegraphics[trim={0cm 0cm 0cm 0.9cm},clip,width=\linewidth]{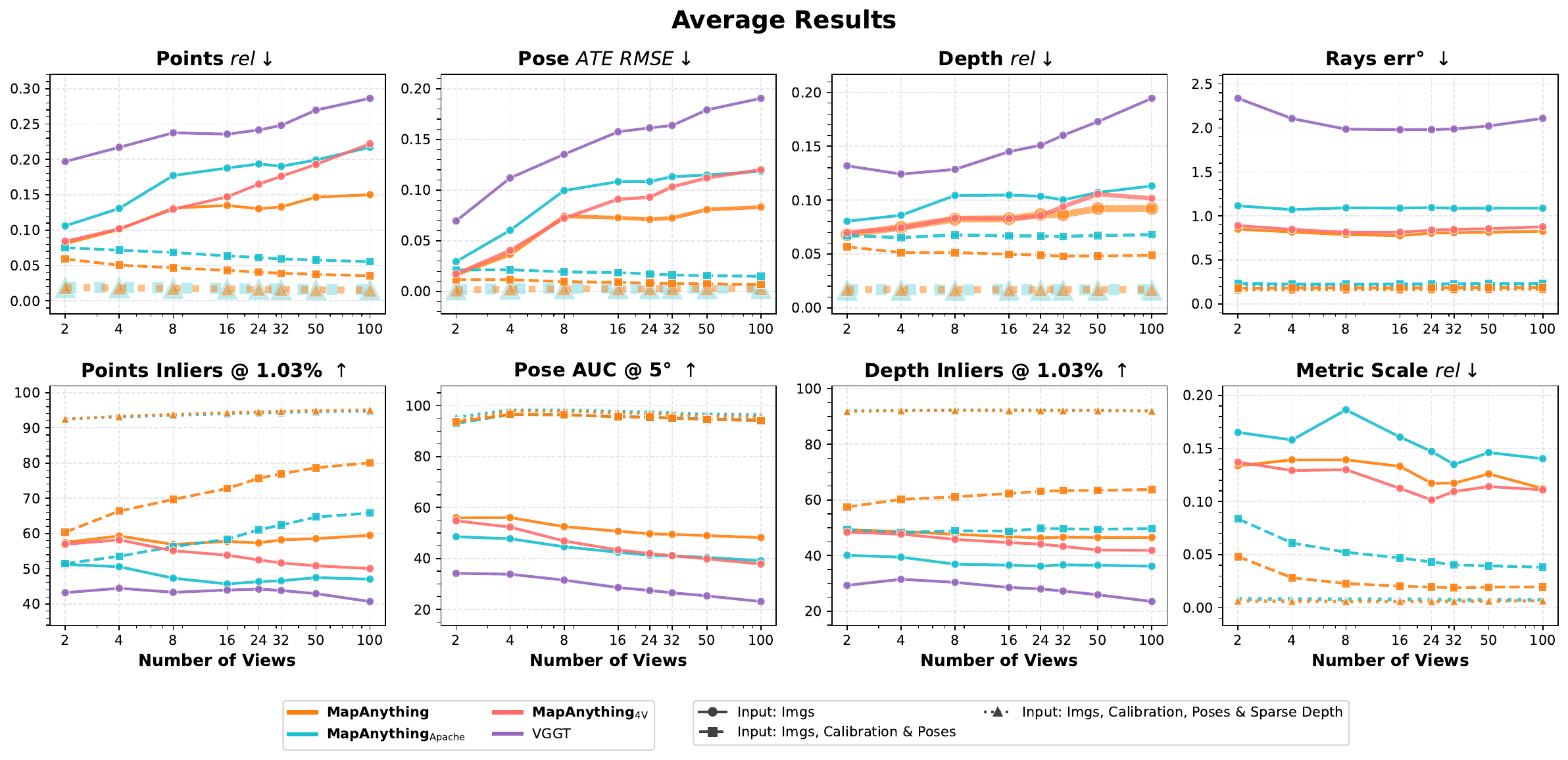}
    \caption{\label{fig:mapa_variants_comparison_graphs}%
    \textbf{The Apache and first stage (up to 4 views) training variants of \coolname show strong dense multi-view reconstruction for input views ranging from 2 to 100 and under different input configurations.}
    We report the absolute relative error ($\absrel$), the inlier ratio at a relative threshold of $1.03\%$ ($\threshI$), the average aligned trajectory error ($ATE~RMSE$), the area under the curve at an error threshold of 5° ($AUC@5$), and the average angular error ($err$) in degrees (°), averaged over ETH3D, ScanNet++ v2 \& TAv2.
    }
\end{figure*}

\begin{figure*}[!t]
    \centering

    \includegraphics[trim={0cm 0cm 0cm 0.9cm},clip,width=\linewidth]{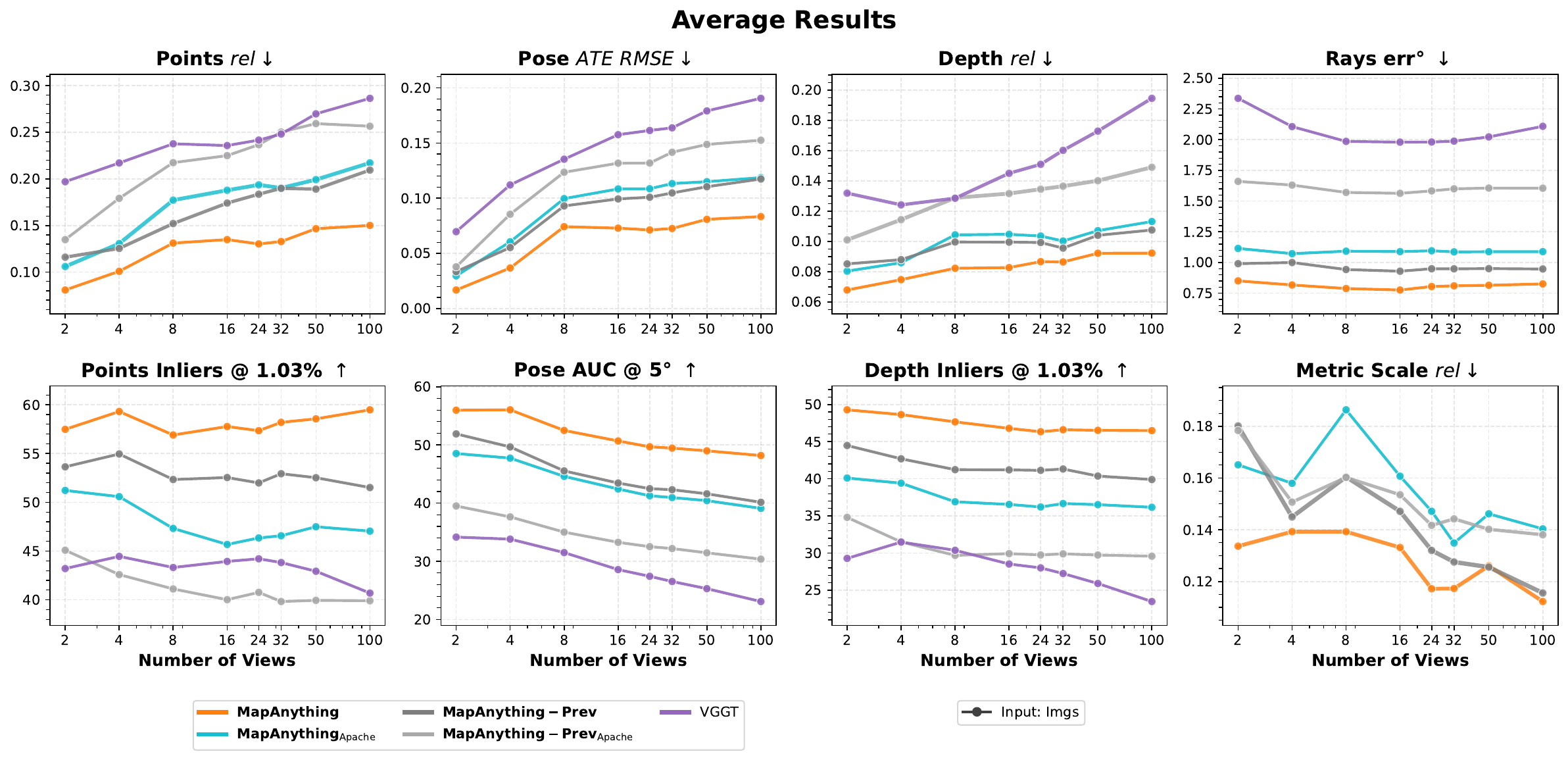}

    \vspace{0.3em}
    
    \includegraphics[trim={0cm 0cm 0cm 0.9cm},clip,width=\linewidth]{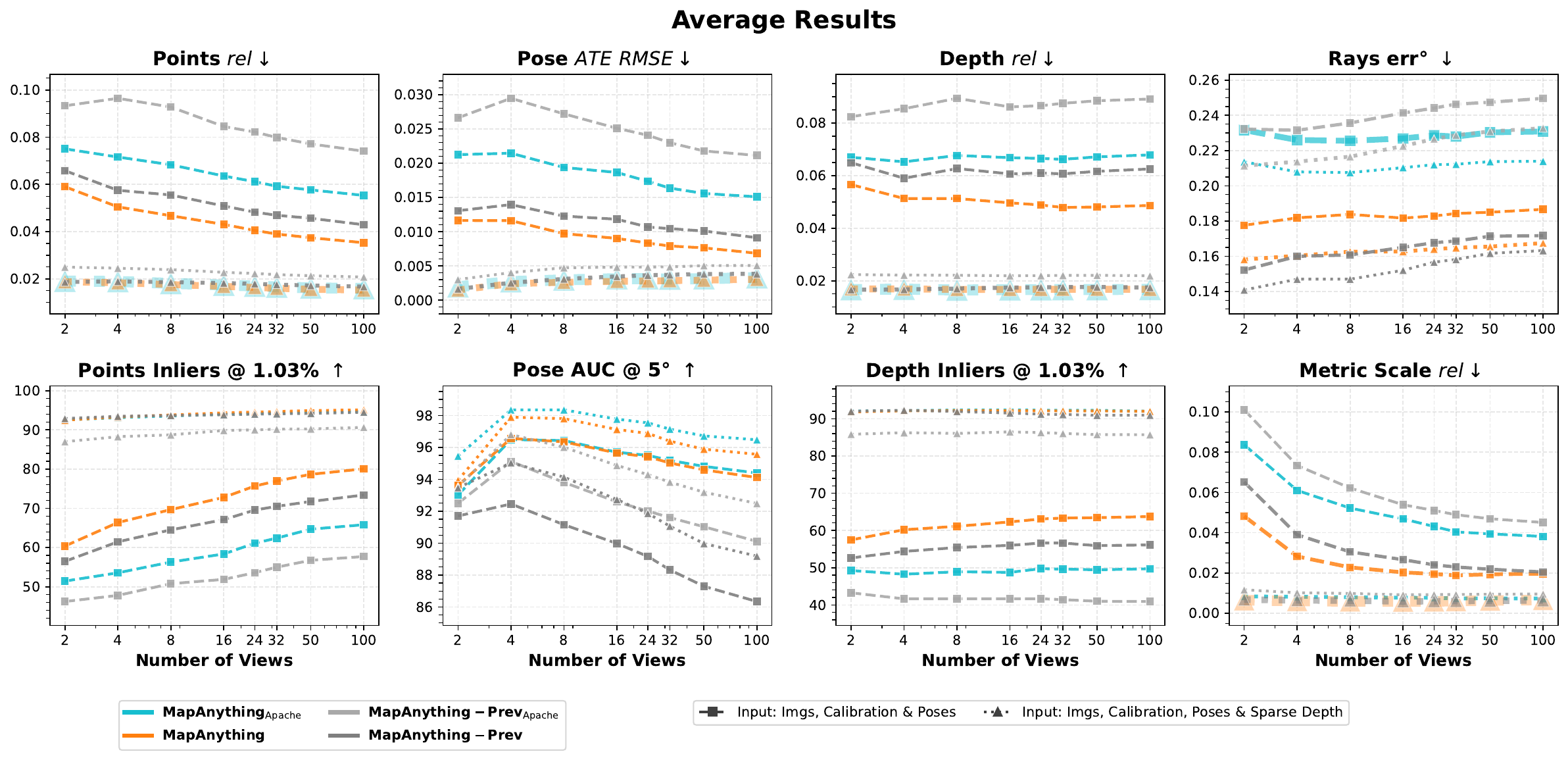}
    
    \caption{\label{fig:v1_1_vs_v1}%
    \textbf{Dense multi-view reconstruction benchmarking comparing the performance of the latest checkpoint as of January 20th 2026 (\coolname) to the one released in September 2025 (\coolname-Prev).}
    The primary difference between the two variants (\textit{latest} \& \textit{previous}) is: (a) Encoder: First 24 layers of 1536-dim DINOv2 ViT-G (\textit{latest}) vs All 24 layers of 1024-dim DINOv2 ViT-L (\textit{previous}), (b) Multi-View Transformer: Last 16 layers of 1536-dim DINOv2 ViT-G (\textit{latest}) vs 24 layers of a randomly initialized 736-dim ViT-B size transformer.
    Both across large-scale trainings and ablation settings, we find that the DINOv2 initialization for the multi-view transformer helps significantly with convergence speed and final performance.
    In the above plots, we report the absolute relative error ($\absrel$), the inlier ratio at a relative threshold of $1.03\%$ ($\threshI$), the average aligned trajectory error ($ATE~RMSE$), the area under the curve at an error threshold of 5° ($AUC@5$), and the average angular error ($err$) in degrees (°), averaged over ETH3D, ScanNet++ v2 \& TAv2.
    }
\end{figure*}

We provide a comparison between different variants of \coolname in \cref{fig:mapa_variants_comparison_graphs}, where the performance of VGGT~\cite{WangCKVRN2025} is provided as a baseline.
First, we show that our two-stage training with covisibility-based view sampling is very effective, where the \coolname model trained for up to 4 views as input already shows strong generalization to a significantly higher number of views.
We also compare the performance of our different open-source models, where, as described in \cref{tab:datasets}, the Apache model is trained on 6 datasets, while the CC-BY-NC one is trained on 13.
While we observe a decrease in performance, the Apache variant is still competitive with the VGGT baseline and its performance further improves as additional geometric inputs are provided.
We also provide a comparison between the various versions of the publicly released models in \cref{fig:v1_1_vs_v1}.

\vspace{-0.15em}

\paragraph{Design Choices:}

As shown in \cref{tab:ablations_loss_attention}, log scaling and alternating attention are key for strong performance when evaluating with 50 views, well beyond the 4 used in training.

\vspace{-0.15em}

\paragraph{Qualitative Examples:}

\begin{figure*}[!ht]
    \centering
    \includegraphics[trim={0cm 8cm 0cm 0cm},clip,width=\linewidth]{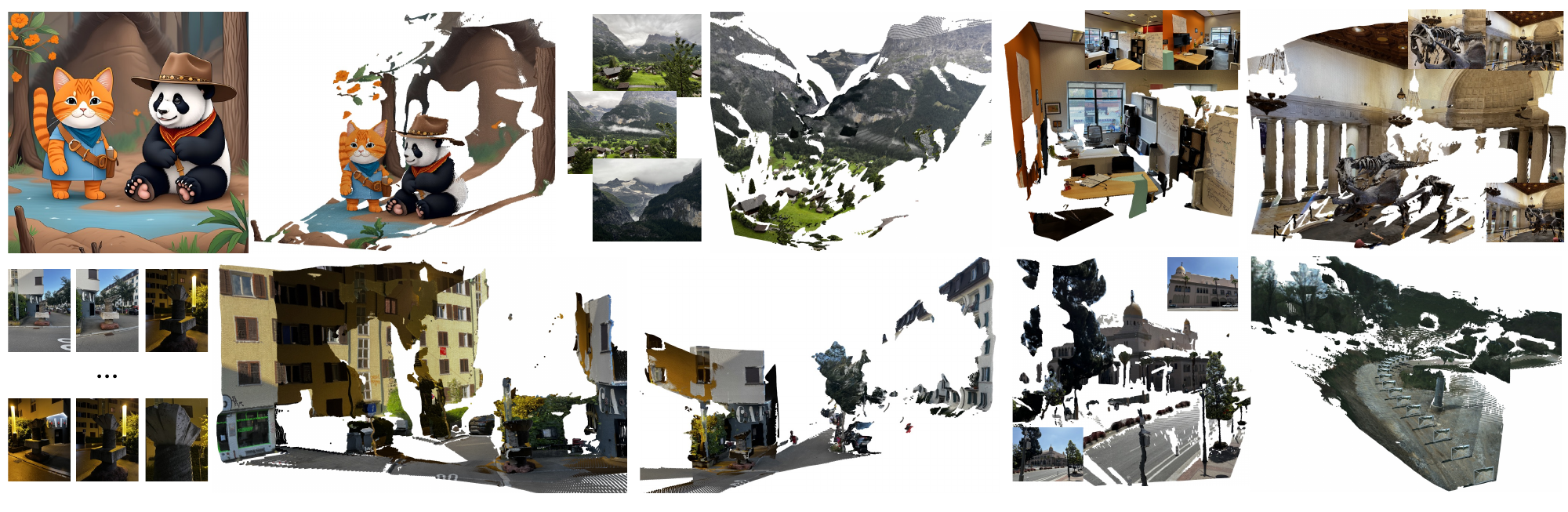}
    \caption{\textbf{\coolname provides high-fidelity dense geometric reconstructions} across varying domains and number of views. While \coolname can support a flexible set of inputs, here we are showcasing its capabilities only using images as input. In particular, we show results across a varying number of captured views and from different domains such as indoor, landscape, art, object-centric, and off-road. It also works well on monocular and art images despite not being trained for it.
}
    \label{fig:qual_diversity}
\end{figure*}

\begin{figure*}[!t]
    \centering
    \includegraphics[width=0.9\linewidth]{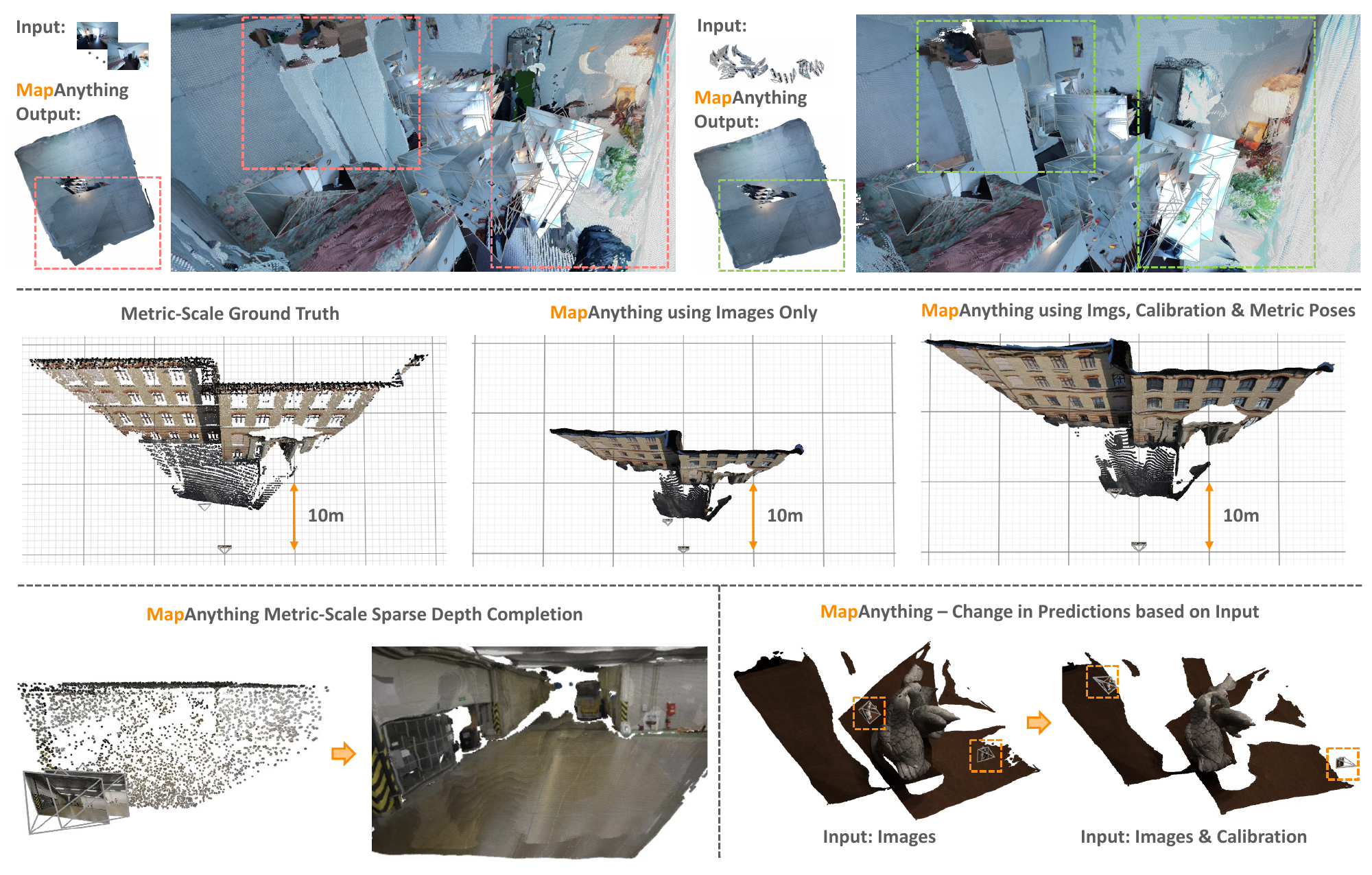}
    \caption{\textbf{Auxiliary geometric inputs improve feed-forward performance of \coolname.} \texttt{(Top)} While \coolname \& other baselines using 100 input images show duplication of 3D structure, when provided with the camera calibration and poses, the 3D reconstruction significantly improves, showing aligned geometry. \texttt{(Middle)} \coolname using images only as input shows non-precise metric scale estimation on ETH3D (a zero-shot dataset). However, when the calibration and metric poses are provided as additional input, the estimated metric scale significantly improves and approximately matches the ground truth. \texttt{(Bottom-Left)} We show that \coolname can leverage a sparse metric point cloud as input to perform dense metric depth completion. \texttt{(Bottom-Right)} Despite not being trained for object-centric data, we show how the scene geometry and cameras change based on the input provided.
    }
    \label{fig:qual_geometric_inputs}
\end{figure*}

\cref{fig:qual_diversity} and \cref{fig:qual_geometric_inputs} showcase the versatility of \coolname.

\vspace{-0.15em}

\paragraph{Comparison with Concurrent Models:}

\begin{figure*}[!t]
    \centering

    \includegraphics[trim={0cm 0cm 0cm 0cm},clip,width=\linewidth]{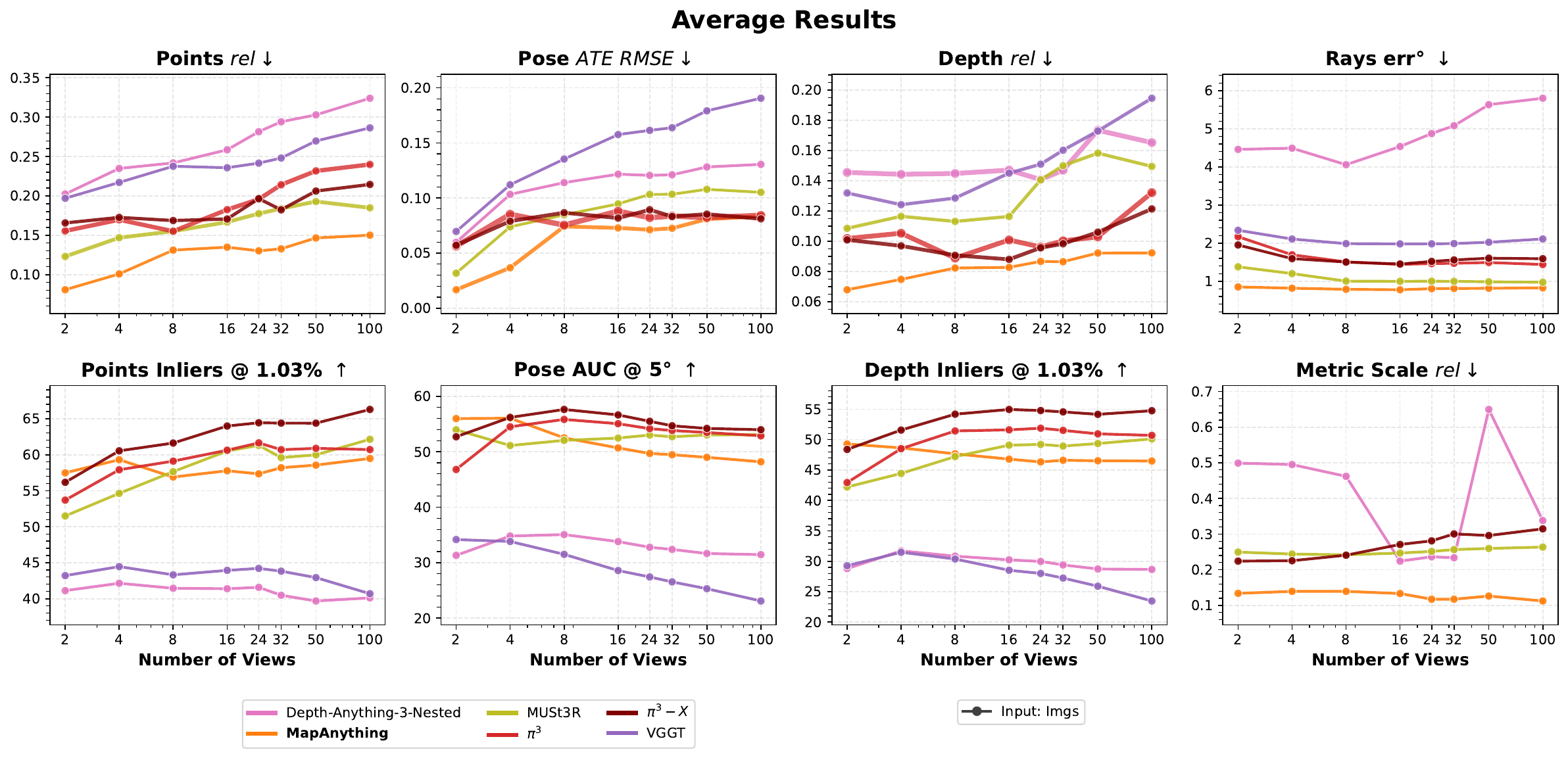}

    \vspace{0.3em}
    
    \includegraphics[trim={0cm 0cm 0cm 0cm},clip,width=\linewidth]{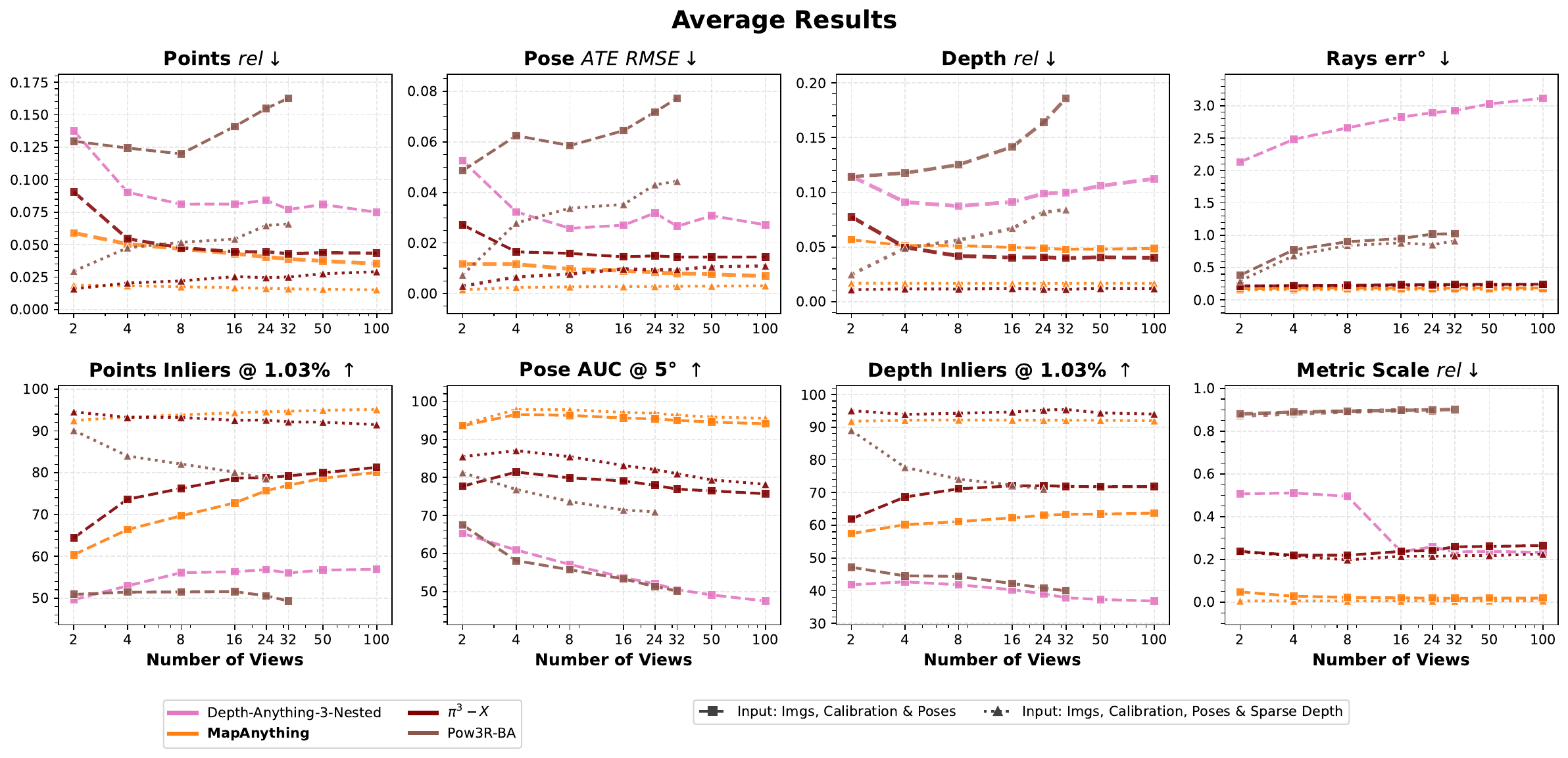}
    
    \caption{\label{fig:v1_1_bench_average}%
    \textbf{\coolname showcases state-of-the-art and competitive dense multi-view reconstruction performance in comparison to latest public concurrent state-of-the-art models as of January 20th 2026, i.e., DA3-Nested \& $\pi^3$-X, despite being trained on significantly less amount of data.}
    We report the absolute relative error ($\absrel$), the inlier ratio at a relative threshold of $1.03\%$ ($\threshI$), the average aligned trajectory error ($ATE~RMSE$), the area under the curve at an error threshold of 5° ($AUC@5$), and the average angular error ($err$) in degrees (°), averaged over ETH3D, ScanNet++ v2 \& TAv2.
    }
\end{figure*}

\begin{figure*}[!t]
    \centering

    \includegraphics[trim={0cm 2cm 0cm 0cm},clip,width=0.9\linewidth]{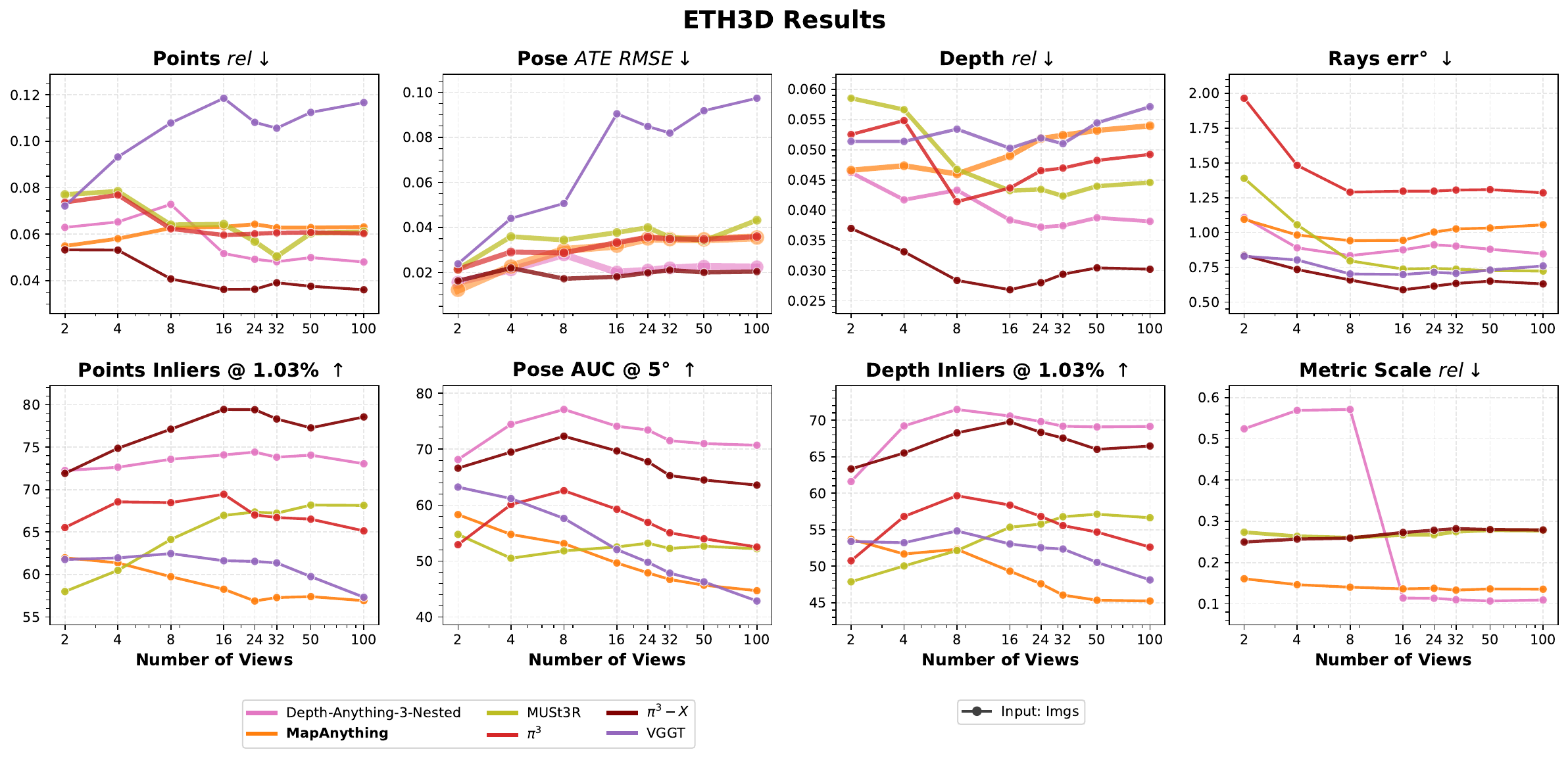}

    \vspace{0.3em}
    
    \includegraphics[trim={0cm 2cm 0cm 0cm},clip,width=0.9\linewidth]{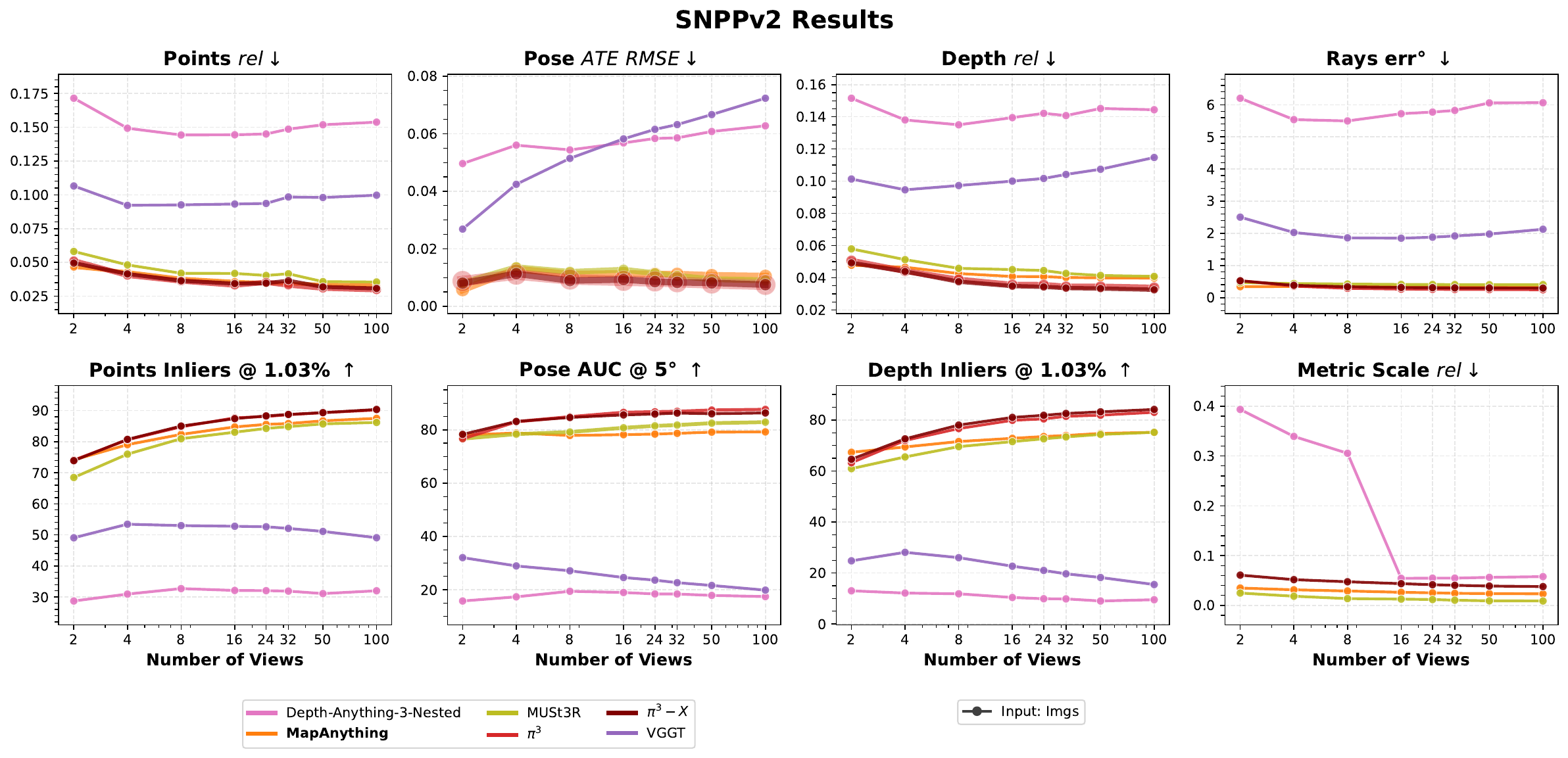}

    \vspace{0.3em}
    
    \includegraphics[trim={0cm 0cm 0cm 0cm},clip,width=0.9\linewidth]{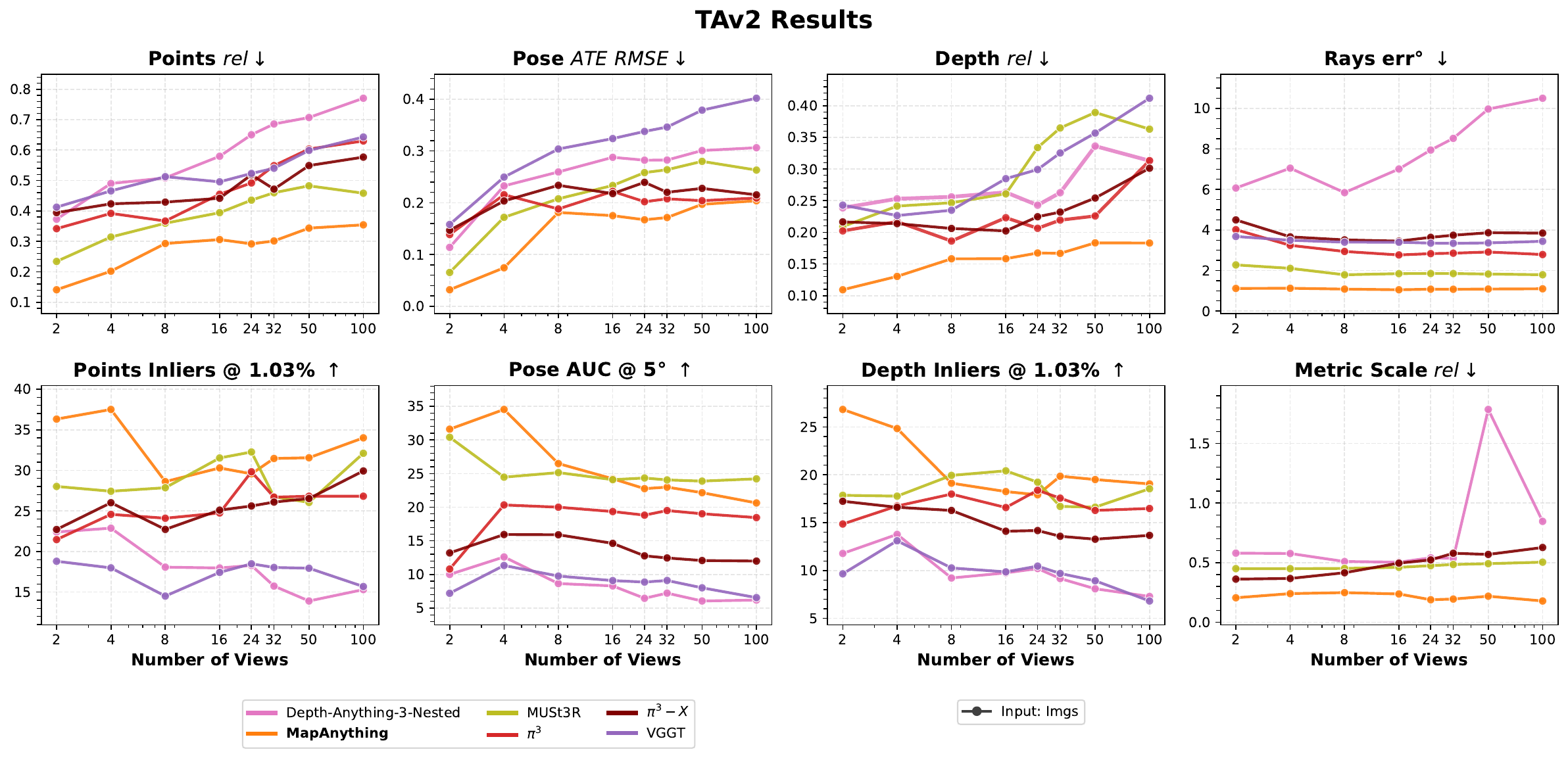}
    
    \caption{\label{fig:v1_1_bench_img_only_per_dataset}%
    \textbf{Dense multi-view reconstruction benchmarking across individual datasets using only images as input.}
    We also include results for latest public concurrent state-of-the-art models as of January 20th 2026.
    Please see \cref{fig:v1_1_bench_average} for details \& the averaged version.
    }
\end{figure*}

\begin{figure*}[!t]
    \centering

    \includegraphics[trim={0cm 2cm 0cm 0cm},clip,width=0.9\linewidth]{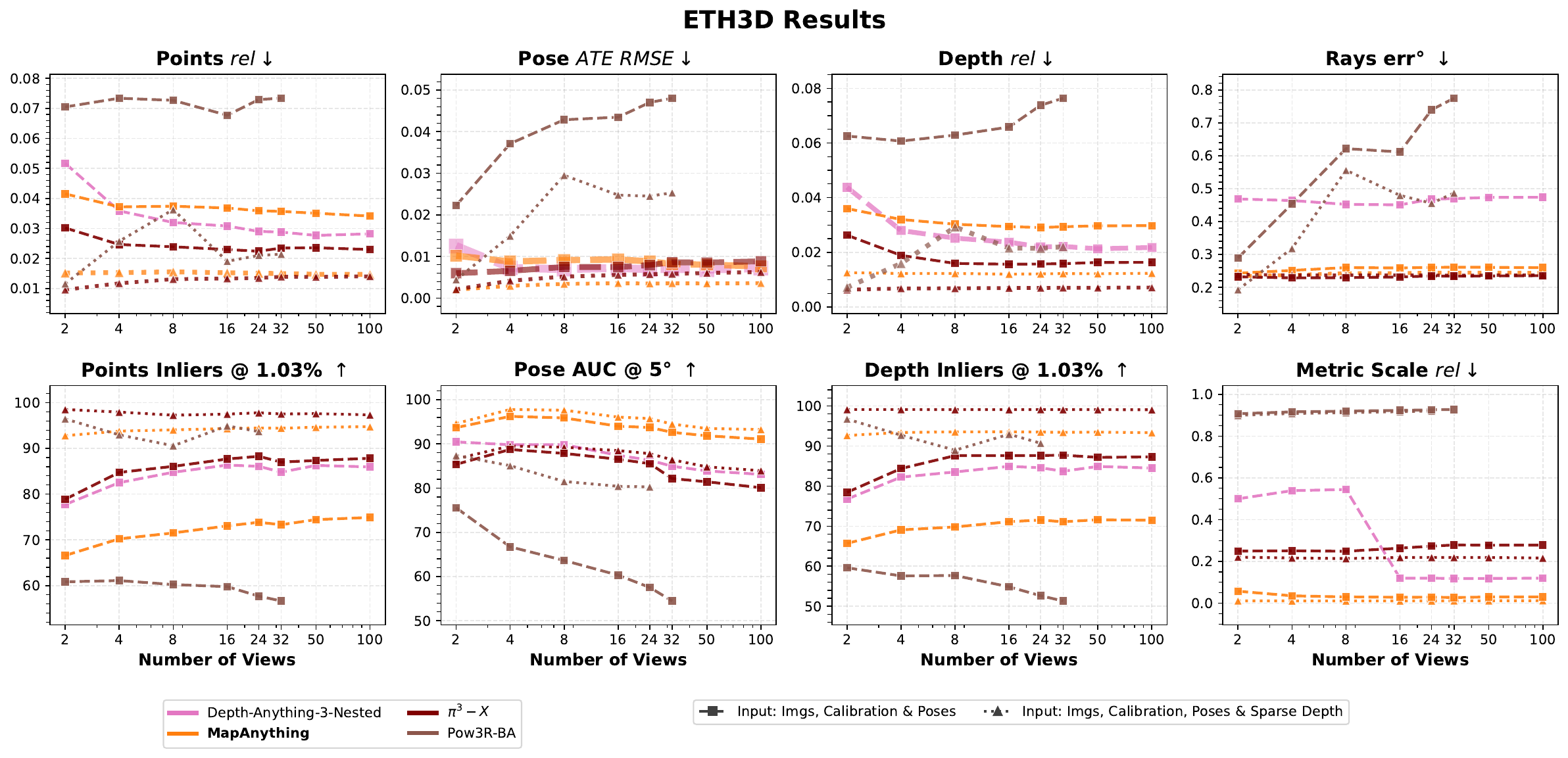}

    \vspace{0.3em}
    
    \includegraphics[trim={0cm 2cm 0cm 0cm},clip,width=0.9\linewidth]{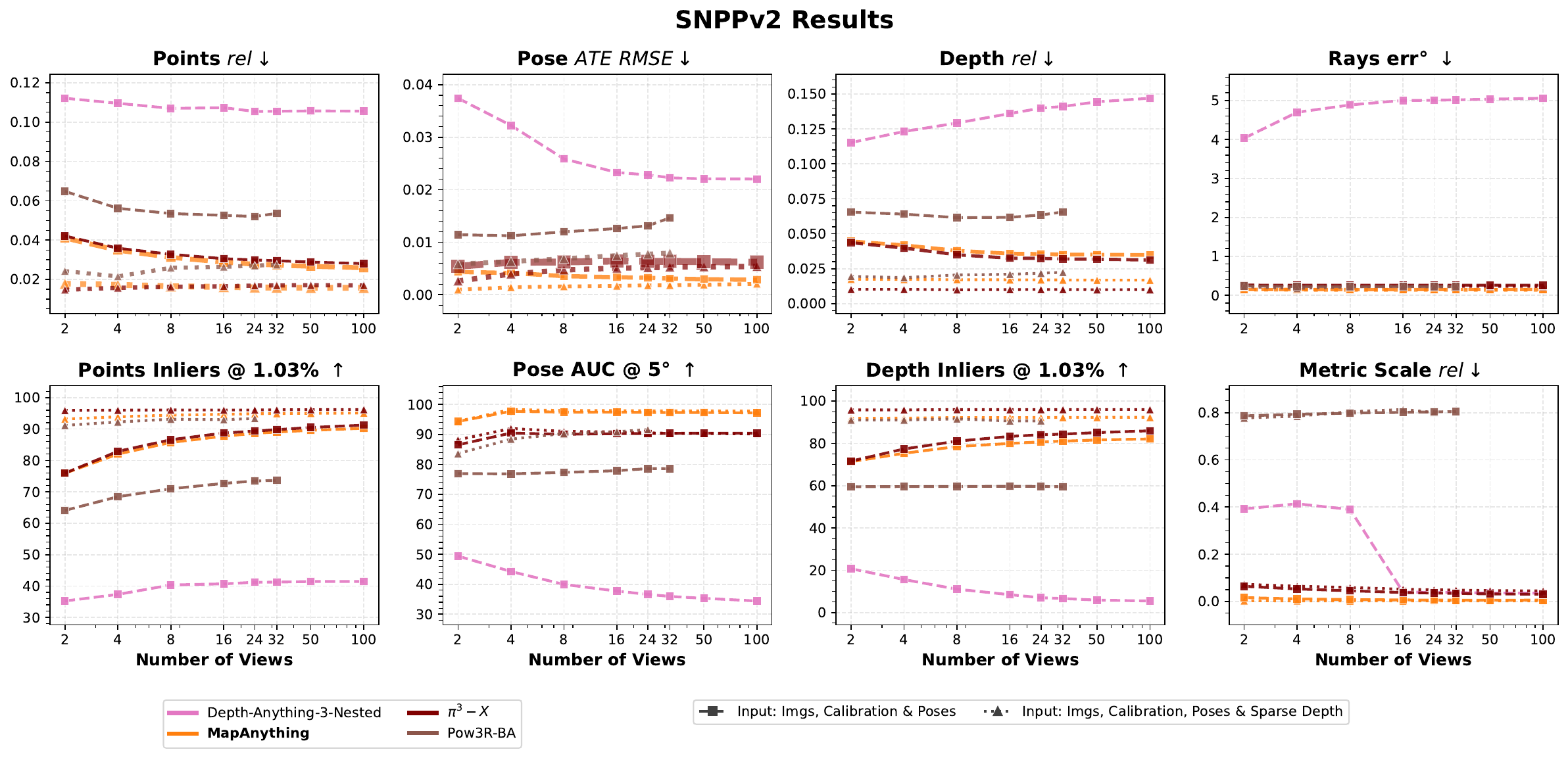}

    \vspace{0.3em}
    
    \includegraphics[trim={0cm 0cm 0cm 0cm},clip,width=0.9\linewidth]{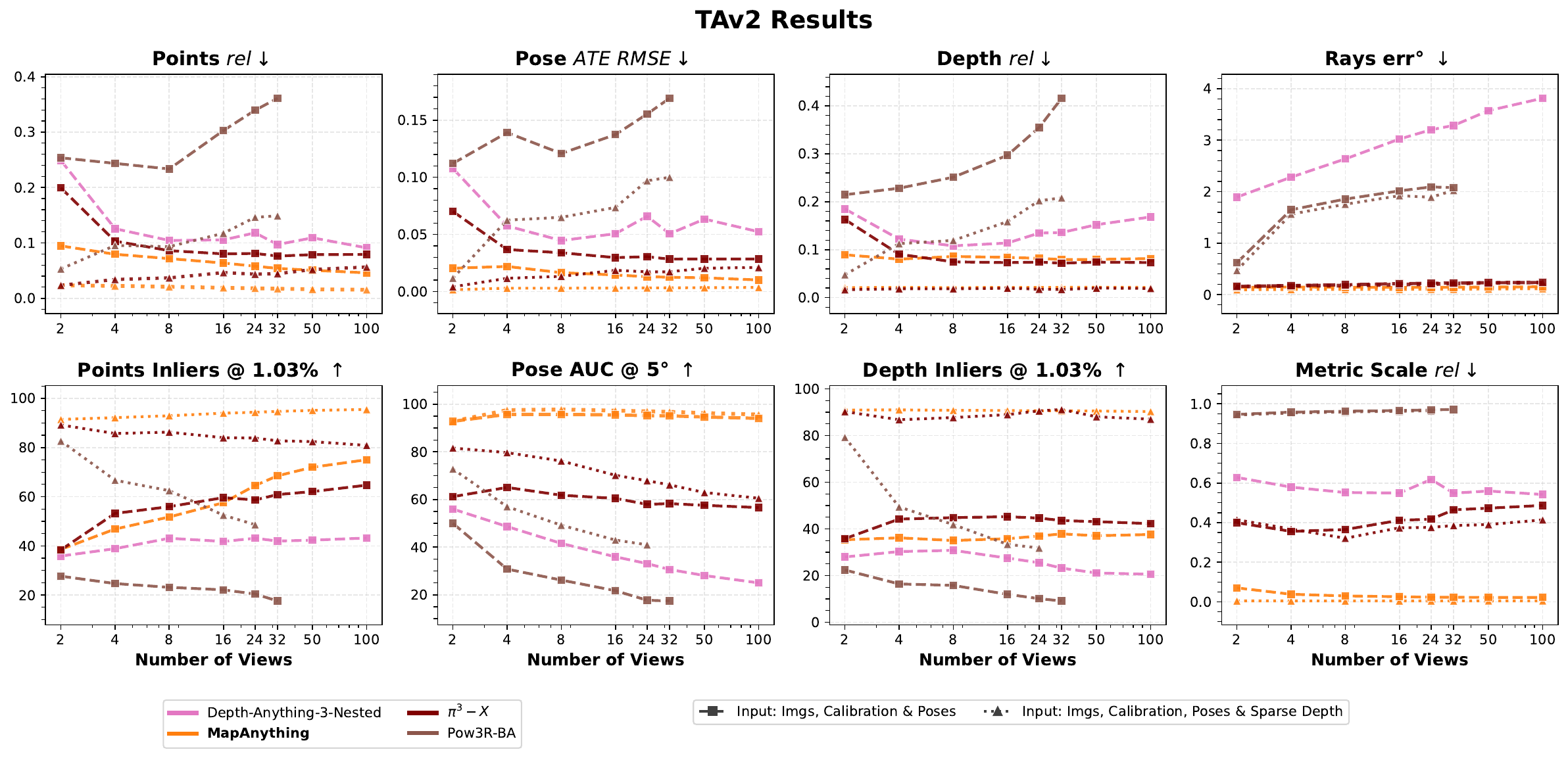}
    
    \caption{\label{fig:v1_1_bench_multi_modal_per_dataset}%
    \textbf{Dense multi-view reconstruction benchmarking across individual datasets using multi-modal inputs.}
    We also include results for latest public concurrent state-of-the-art models as of January 20th 2026.
    Please see \cref{fig:v1_1_bench_average} for details \& the averaged version.
    }
\end{figure*}

\cref{fig:v1_1_bench_average} showcases the average results, while \cref{fig:v1_1_bench_img_only_per_dataset} and \cref{fig:v1_1_bench_multi_modal_per_dataset} show the performance for each individual dataset.

\clearpage

\clearpage

\begin{table*}[t]
    \centering
    \begin{minipage}{0.48\textwidth}
        \centering

\input{tables/mapanything_input_variants}

    \end{minipage}
    \hfill %
    \begin{minipage}{0.48\textwidth}
        \centering
        \input{tables/ablations_loss_attention}
    \end{minipage}
\end{table*}

\clearpage

\clearpage

\clearpage

\clearpage

\clearpage

%% file: text/06_acknowledgments.tex
\section{Acknowledgments}

We thank Michael Zollhöfer for his initial involvement in project discussions.
We also thank Jeff Tan, Jianyuan Wang, Jay Karhade, Jinghao (Jensen) Zhou, Yifei Liu, Shubham Tulsiani, Khiem Vuong, Yuheng Qiu, Shibo Zhao, Omar Alama, Andrea Simonelli, Corinne Stucker, Denis Rozumny, Bardienus Duisterhof, and Wenshan Wang for their insightful discussions, feedback, and assistance with parts of the project.
Lastly, we appreciate the support for compute infrastructure from Julio Gallegos, Tahaa Karim, and Ali Ganjei.

\vspace{0.4em}

\noindent\textbf{\textcolor{citegray}{Funding at Carnegie Mellon University:}} Nikhil's time and parts of this work at CMU was supported by Defense Science and Technology Agency (DSTA) Contract {\scriptsize\texttt{\#DST000EC124000205}}, DEVCOM Army Research Laboratory (ARL) SARA Degraded SLAM {\scriptsize\texttt{CRA W911NF-20-S-0005}}, and the Intelligence Advanced Research Projects Activity (IARPA) via Department of Interior/Interior Business Center (DOI/IBC) contract number {\scriptsize\texttt{140D0423C0074}}. The U.S. Government is authorized to reproduce and distribute reprints for Governmental purposes notwithstanding any copyright annotation thereon. Disclaimer: The views and conclusions contained herein are those of the authors and should not be interpreted as necessarily representing the official policies or endorsements, either expressed or implied, of IARPA, DOI/IBC, or the U.S. Government. 
The compute required for this work at CMU was supported by a hardware grant from Nvidia and used PSC Bridges-2 through allocation {\scriptsize\texttt{cis220039p}} from the Advanced Cyberinfrastructure Coordination Ecosystem: Services \& Support (ACCESS) program.

%% file: tables/mapanything_input_variants.tex
\caption{\label{tab:supplementary_mapanything_input_variants}%
  \textbf{\coolname demonstrates remarkable flexibility in handling diverse input configurations, with performance improving as additional modalities are provided.}
  While our universal training supports 64 (i.e., $2^6$) possible input combinations, we highlight 12 representative combinations in this table.
  We report the absolute relative error ($\absrel$) and the inlier ratio at a relative threshold of 1.03\% ($\threshI$) for 50 views, averaged over ETH3D, ScanNet++ v2 \& TAv2.
  `K' denotes camera intrinsics and `sparse' depth indicates that 90\% of the valid depth is randomly masked out.
}
\centering
\scriptsize
\setlength{\tabcolsep}{1.5mm}
\begin{tabular}{cccccccccc}
\toprule
\multicolumn{7}{c}{\textbf{\coolname Inputs}}
& \multicolumn{3}{c}{\textbf{Avg. Performance}} \\
\cmidrule(lr){1-7} \cmidrule(lr){8-10}
& & & \multicolumn{2}{c}{\textbf{Depth}} & \multicolumn{2}{c}{\textbf{Metric Scale}}
& \textbf{Scale}
& \multicolumn{2}{c}{\textbf{Points}} \\
\textbf{Imgs} & \textbf{K} & \textbf{Poses}
& \textbf{Dense} & \textbf{Sparse} & \textbf{Pose} & \textbf{Depth}
& $\absrel\downarrow$
& $\absrel\downarrow$ & $\threshI\uparrow$
\\
\midrule
\multicolumn{10}{l}{\textbf{a) Images Only}} \\[0.4em]
\greencheck & \redx & \redx & \redx & \redx & \redx & \redx         
  & 0.13   
  & 0.15           
  & 58.6       
  \\

\arrayrulecolor{lightgray}\midrule
\multicolumn{10}{l}{\textbf{b) Images \& Intrinsics}} \\[0.4em]
\greencheck & \greencheck & \redx & \redx & \redx & \redx & \redx         
  & 0.13   
  & 0.14           
  & 61.5       
  \\

\arrayrulecolor{lightgray}\midrule
\multicolumn{10}{l}{\textbf{c) Images \& Poses}} \\[0.4em]
\greencheck & \redx & \greencheck & \redx & \redx & \redx & \redx         
  & 0.11   
  & 0.04           
  & 76.8       
  \\
\greencheck & \redx & \greencheck & \redx & \redx & \greencheck & \redx         
  & 0.02   
  & 0.04           
  & 75.3       
  \\

\arrayrulecolor{lightgray}\midrule
\multicolumn{10}{l}{\textbf{d) Images, Intrinsics \& Poses}} \\[0.4em]
\greencheck & \greencheck & \greencheck & \redx & \redx & \redx & \redx         
  & 0.11   
  & 0.03           
  & 80.5       
  \\
\greencheck & \greencheck & \greencheck & \redx & \redx & \greencheck & \redx         
  & 0.02   
  & 0.04           
  & 78.7       
  \\

\arrayrulecolor{lightgray}\midrule
\multicolumn{10}{l}{\textbf{e) Images \& Depth}} \\[0.4em]
\greencheck & \greencheck & \redx & \redx & \greencheck & \redx & \greencheck         
  & 0.04   
  & 0.11           
  & 77.2       
  \\
\greencheck & \greencheck & \redx & \greencheck & \redx & \redx & \greencheck         
  & 0.05   
  & 0.11           
  & 67.8       
  \\

\arrayrulecolor{lightgray}\midrule
\multicolumn{10}{l}{\textbf{f) Images, Intrinsics, Poses \& Depth}} \\[0.4em]
\greencheck & \greencheck & \greencheck & \redx & \greencheck & \redx & \redx         
  & 0.12   
  & 0.02           
  & 92.0       
  \\
\greencheck & \greencheck & \greencheck & \greencheck & \redx & \redx & \redx         
  & 0.12   
  & 0.02           
  & 83.0       
  \\
\greencheck & \greencheck & \greencheck & \redx & \greencheck & \greencheck & \greencheck         
  & 0.01   
  & 0.02           
  & 94.9       
  \\
\greencheck & \greencheck & \greencheck & \greencheck & \redx & \greencheck & \greencheck         
  & 0.01   
  & 0.02           
  & 84.4       
  \\

\arrayrulecolor{black}\bottomrule
\end{tabular}

%% file: tables/ablations_loss_attention.tex
  \centering
  \caption{\label{tab:ablations_loss_attention}%
    \textbf{Ablations showing loss and multi-view transformer attention design choices critical for strong reconstruction performance.} %
    We report the absolute relative error ($\absrel$) and the inlier ratio at a relative threshold of 1.03\% ($\threshI$) at 50 views, averaged over ETH3D, ScanNet++ v2 \& TAv2.
    Best results are \textbf{bold}.
  }
  \begin{subtable}[t]{0.49\linewidth}
    \centering
    \caption{\textbf{Loss Scheme}}
    \input{tables/ablation_loss}
    \label{tab:ablation_loss}
  \end{subtable}%
  \hfill
  \begin{subtable}[t]{0.50\linewidth}
    \centering
    \caption{\textbf{Attention Scheme}}
    \input{tables/ablation_attention}
    \label{tab:ablation_attention}
  \end{subtable}%

%% file: tables/ablation_loss.tex
\resizebox{\textwidth}{!}{
\begin{tabular}{l|c|cc|}
\toprule
    & \multicolumn{3}{c|}{\textbf{ETH3D, SN++v2 \& TAV2}}

    \\

    \cmidrule(lr){2-4} 

    & \textbf{Metric Scale} & \multicolumn{2}{c|}{\textbf{Pointmaps}}

    \\

    \cmidrule(lr){2-2} \cmidrule(lr){3-4}

    \textbf{Methods}
    & $\absrel\downarrow$
    & $\absrel\downarrow$ & $\threshI\uparrow$

    \\

    \midrule

    \multicolumn{4}{l}{\textbf{Input: Images Only}}
    \\

    \emph{Overall Factored Loss}
    & \textbf{0.16}
    & \textbf{0.29}
    & \textbf{31.8}
    \\

    No Log Loss
    & 0.17
    & 0.39
    & 27.3
    \\

\bottomrule
\end{tabular}
}

%% file: tables/ablation_attention.tex
\resizebox{\textwidth}{!}{
\begin{tabular}{l|c|cc|}
\toprule
    & \multicolumn{3}{c|}{\textbf{ETH3D, SN++v2 \& TAV2}}

    \\

    \cmidrule(lr){2-4} 

    & \textbf{Metric Scale} & \multicolumn{2}{c|}{\textbf{Pointmaps}}

    \\

    \cmidrule(lr){2-2} \cmidrule(lr){3-4}

    \textbf{Methods}
    & $\absrel\downarrow$
    & $\absrel\downarrow$ & $\threshI\uparrow$

    \\

    \midrule

    \multicolumn{4}{l}{\textbf{Input: Images Only}}
    \\

    \emph{Alternating}~\cite{WangCKVRN2025}
    & \textbf{0.16}
    & \textbf{0.29}
    & \textbf{31.8}
    \\

    Global w/ View PE~\cite{YangSLHTCCMF2025}
    & 0.20
    & 0.53
    & 19.7
    \\

\bottomrule
\end{tabular}
}